\documentclass[wcp]{jmlr}

\usepackage[utf8]{inputenc} 
\usepackage[T1]{fontenc}    
\usepackage{booktabs}
\usepackage{hyperref}       
\usepackage{url}            
\usepackage{amsfonts}       
\usepackage{nicefrac}
\usepackage{subcaption}
\usepackage{float}
\usepackage{graphicx}
\usepackage{tikz}
\usepackage{algorithm}
\usepackage{algpseudocode}
\usepackage{ntheorem}

\usetikzlibrary{arrows}
\usetikzlibrary{shapes.misc}
\usetikzlibrary{positioning}
\usetikzlibrary{decorations.pathreplacing}
\tikzset{>=latex}

\usepackage{longtable}

\usepackage{booktabs}

\newcommand\ddfrac[2]{\frac{\displaystyle #1}{\displaystyle #2}}

\jmlrvolume{80}
\jmlryear{2017}
\jmlrworkshop{ACML 2017}

\title[Accumulated Gradient Normalization]{Accumulated Gradient Normalization}

\author{\Name{Joeri R. Hermans}\Email{joeri.hermans@doct.ulg.ac.be}\\
  \Name{Gerasimos Spanakis} \Email{jerry.spanakis@maastrichtuniversity.nl}\\
  \Name{Rico M\"ockel} \Email{rico.mockel@maastrichtuniversity.nl}\\
  \addr Department of Electrical Engineering and Computer Science, Li\`ege University, Belgium\\
  \addr Department of Data Science \& Knowledge Engineering, Maastricht University, The Netherlands
}

\begin{document}

\maketitle

\begin{abstract}
This work addresses the instability in asynchronous data parallel optimization. It does so by introducing a novel distributed optimizer which is able to efficiently optimize a centralized model under communication constraints. The optimizer achieves this by pushing a normalized sequence of first-order gradients to a parameter server. This implies that the magnitude of a worker delta is smaller compared to an accumulated gradient, and provides a better direction towards a minimum compared to first-order gradients, which in turn also forces possible implicit momentum fluctuations to be more aligned since we make the assumption that all workers contribute towards a single minima. As a result, our approach mitigates the parameter staleness problem more effectively since staleness in asynchrony induces (implicit) momentum, and achieves a better convergence rate compared to other optimizers such as asynchronous \textsc{easgd} and \textsc{dynsgd}, which we show empirically.
\end{abstract}
\begin{keywords}
Distributed Optimization, Neural Networks, Gradient Descent
\end{keywords}

\section{Introduction}
\label{sec:introduction}

Speeding up gradient based methods has been a subject of interest over the past years with many practical applications, especially with respect to Deep Learning. Despite the fact that many optimizations have been done on a hardware level, the convergence rate of very large models remains problematic. Therefore, data parallel methods next to mini-batch parallelism have been suggested~\cite{dean2012large, ho2013more, hadjis2016omnivore, recht2011hogwild, louppe2010zealous, jiang2017heterogeneity, zhang2015deep} to further decrease the training time of parameterized models using gradient based methods. Nevertheless, asynchronous optimization was considered too unstable for practical purposes due to a lacking understanding of the underlying mechanisms, which is an issue this work addresses.\\

Data Parallelism is an inherently different methodology of optimizing parameters. As stated above, it is a technique to reduce the overall training time of a model. In essence, data parallelism achieves this by having $n$ workers optimizing a central model, and at the same time, processing $n$ different shards (partitions) of the dataset in parallel over multiple workers\footnote{A worker in this work is a process on a single machine. However, it is possible that multiple workers share the same machine. Nevertheless, one could construct the distribution mechanism (even manually) in such a way every worker will be placed on a different machine.}. The workers are coordinated in such a way that they optimize the parameterization of a central model or central variable, which we denote by $\tilde{\theta}_t$. The coordination mechanism of the workers can be implemented in many different ways. A popular approach to coordinate workers is to employ a centralized \emph{Parameter Server} (PS). The sole responsibility of the parameter server is to aggregate model updates coming from the workers (\emph{worker commits}), and to handle parameter requests (\emph{worker pulls}).\\

Recently, a theoretical contribution has been made~\cite{implicitmomentum} which defines asynchronous optimization in terms of (implicit) \emph{momentum} due to the presence of a queuing model of gradients based on past parameterizations. This paper mainly builds upon this work, and~\cite{zhang2015deep} to construct a better understanding why asynchronous optimization shows proportionally more divergent behavior when the number of parallel workers increases, and how this affects existing distributed optimization algorithms.

\begin{figure}[H]
  \centering
  \begin{tikzpicture}[scale=0.9, every node/.style={scale=0.9}]
    \node at (-5.5, 0) {$t$};
    \draw[->] (-5, 0) -- (5, 0);

    \draw[->] (-4, 0) -- (-4, -1);
    \draw[->] (-3.1, 0) -- (-3.1, -1);
    \draw[rounded corners=2pt] (-4.4, -1) rectangle (-3.6, -1.5) node[pos=.55] {$w_1$};
    \draw[rounded corners=2pt] (-3.5, -1) rectangle (-2.7, -1.5) node[pos=.55] {$w_2$};
    \draw [decorate,decoration={brace,amplitude=5pt}]
    (-4.1, 0.1) -- (-3, 0.1) node [black,midway,yshift=13pt] {\small $\tilde{\theta}_t$};

    \draw[->] (-1, 0) --(-1, -1);
    \draw[<-] (-1.2, 0) -- (-1.2, -1);
    \node at (-2.1, -0.6) {\small $\nabla_\theta \mathcal{L}_1(\tilde{\theta}_t)$};
    \draw[rounded corners=2pt] (-1.1 - 0.4, -1) rectangle (-1.1 + 0.4, -1.5) node[pos=.55] {$w_1$};
    \node at (-1.1, 0.5) {\small $\tilde{\theta}_{t+1} = \tilde{\theta}_t - \eta_t \odot \nabla_\theta \mathcal{L}_1(\tilde{\theta}_t)$};

    \draw[<-] (3, 0) --(3, -1);
    \draw[->] (3.2, 0) -- (3.2, -1);
    \node at (2.1, -0.6) {\small $\nabla_\theta \mathcal{L}_2(\tilde{\theta}_t)$};
    \draw[rounded corners=2pt] (3.1 - 0.4, -1) rectangle (3.1 + 0.4, -1.5) node[pos=.55] {$w_2$};
    \node at (3.1, 0.5) {\small $\tilde{\theta}_{t+2} = \tilde{\theta}_{t+1} - \eta_t \odot \nabla_\theta \mathcal{L}_2(\tilde{\theta}_t)$};

    \node at (-2, 2) {Updating $\tilde{\theta}_{t+1}$ with gradient $\nabla_\theta \mathcal{L}_2(\tilde{\theta}_t)$ based on $\tilde{\theta}_t$};
    \draw (-2.6, 1.7) edge[out=-30, in=110, ->] (2.4, 0.8);
    \draw (2.2, 1.95) edge[out=0, in=110, ->] (4.15, 0.8);
  \end{tikzpicture}
  \caption{In Asynchronous Data Parallelism workers compute and commit gradients to the PS asynchronously. This has as a side-effect that some workers are computing, and thus committing, gradients based on old values. These gradients are called \emph{stale gradients} in literature. In this particular example there are 2 workers $w_1$, and $w_2$. At the start of the optimization process, both workers pull the most recent parameterization, $\tilde{\theta}_t$, from the PS. Now all workers start computing gradients asynchronously based on the pulled parameterization. However, since the PS incorporates gradients into the center variable asynchronously as a simple queuing (FIFO) model, other workers will update the center variable with gradients based on a stale parameterization, as shown above. Finally, assuming that the computing cluster is homogeneous, we can derive from this figure that the expected staleness of a gradient update is $\mathbf{E}[\tau] = (n - 1)$, as mentioned in~\cite{implicitmomentum}.}
  \label{fig:intro_asyn_data_parallelism}
\end{figure}
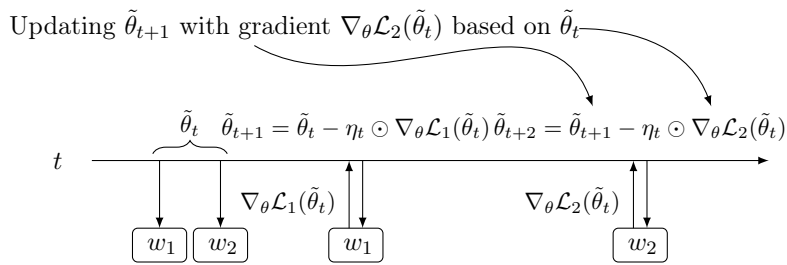

The rest of the paper is organized as follows. In Section~\ref{sec:concept_and_intuition} we present the intuition for our method. Section~\ref{sec:method} describes the proposed method in full, followed by an experimental validation in Section~\ref{sec:experiments}. Finally, Section~\ref{sec:conclusion} concludes the paper by giving an overview of the contributions presented in this work. 

\section{Concept \& Intuition}
\label{sec:concept_and_intuition}

\begin{figure}
    \centering
    \subfigure[$n = 10$]{\label{fig:downpour_convergence_a}\includegraphics[width=.45\linewidth]{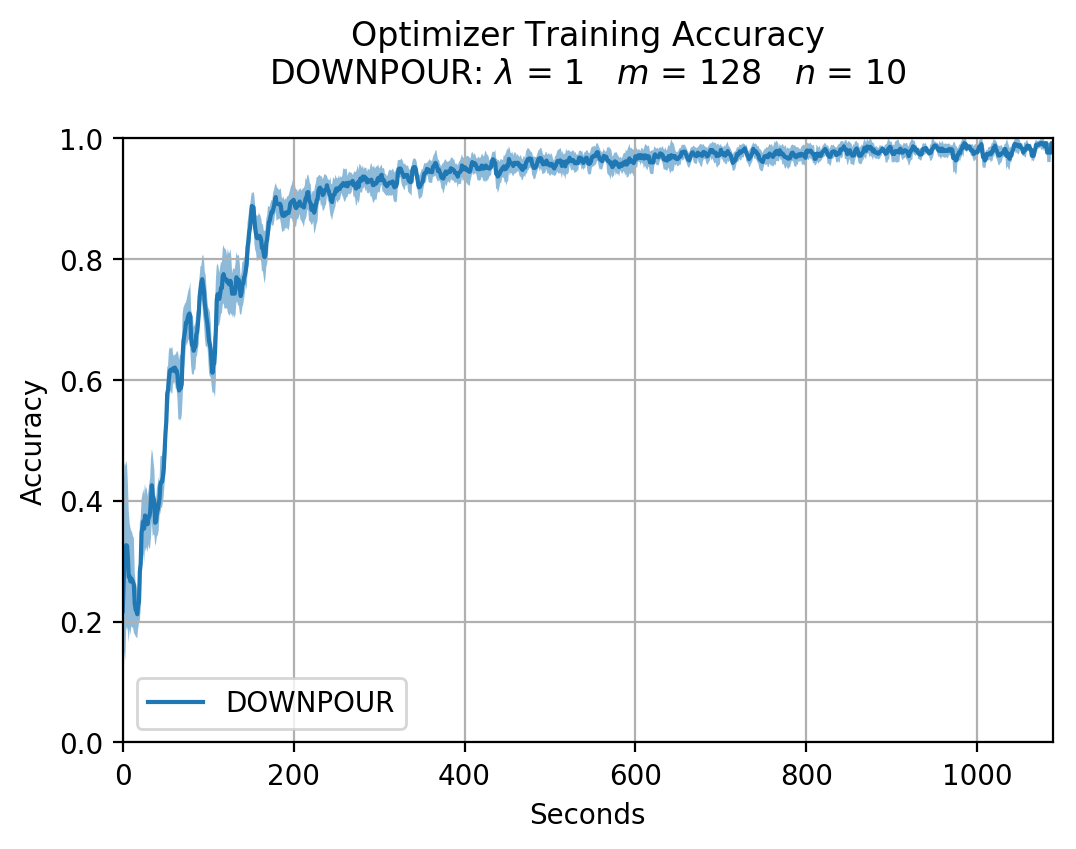}}
    \subfigure[$n = 20$]{\label{fig:downpour_convergence_b}\includegraphics[width=.45\linewidth]{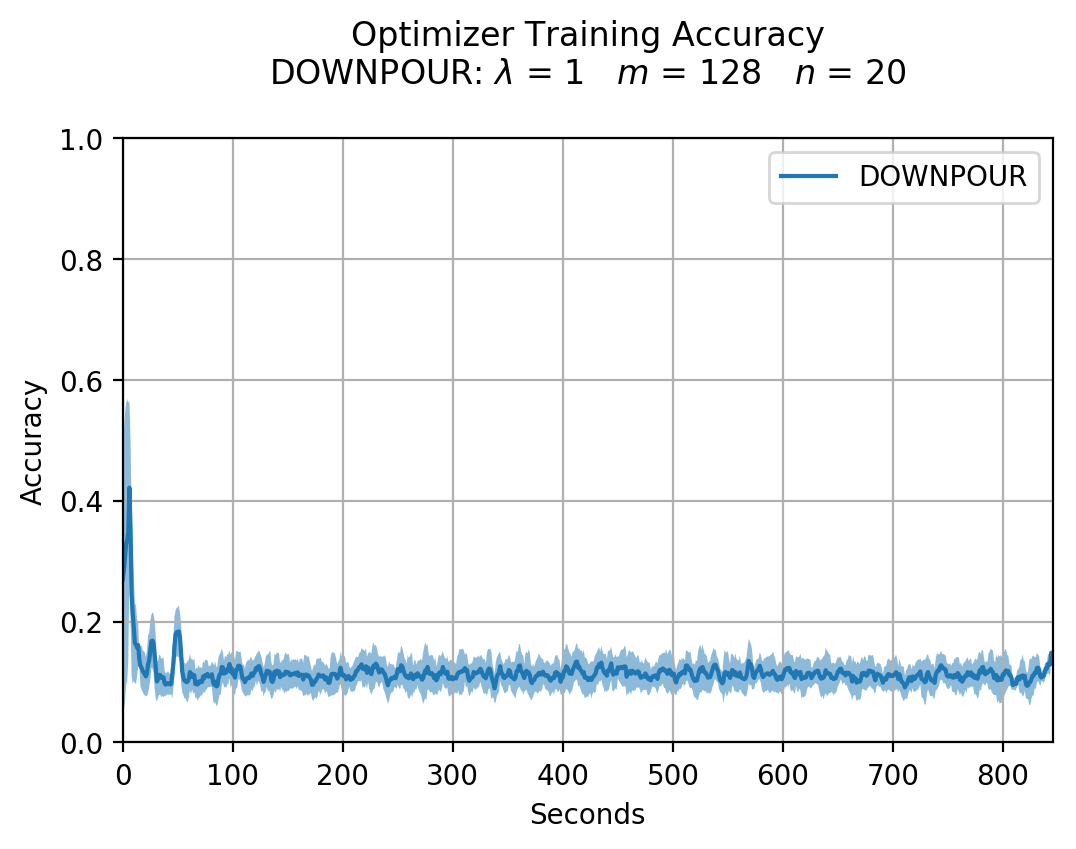}}
    \caption{\textsc{downpour} divergence due to number ($n = 20$) of asynchronous workers in the optimization process~\cite{implicitmomentum} for this particular problem, and not dealing with parameter staleness in a more intelligent way. Lowering the number workers ($n = 10$) causes the central variable to converge.}
    \label{fig:downpour_convergence}
\end{figure}

\begin{figure}
  \centering
  \subfigure[No Gradient Accumulation]{\label{fig:agn_intuition_a}\includegraphics[width=.45\textwidth]{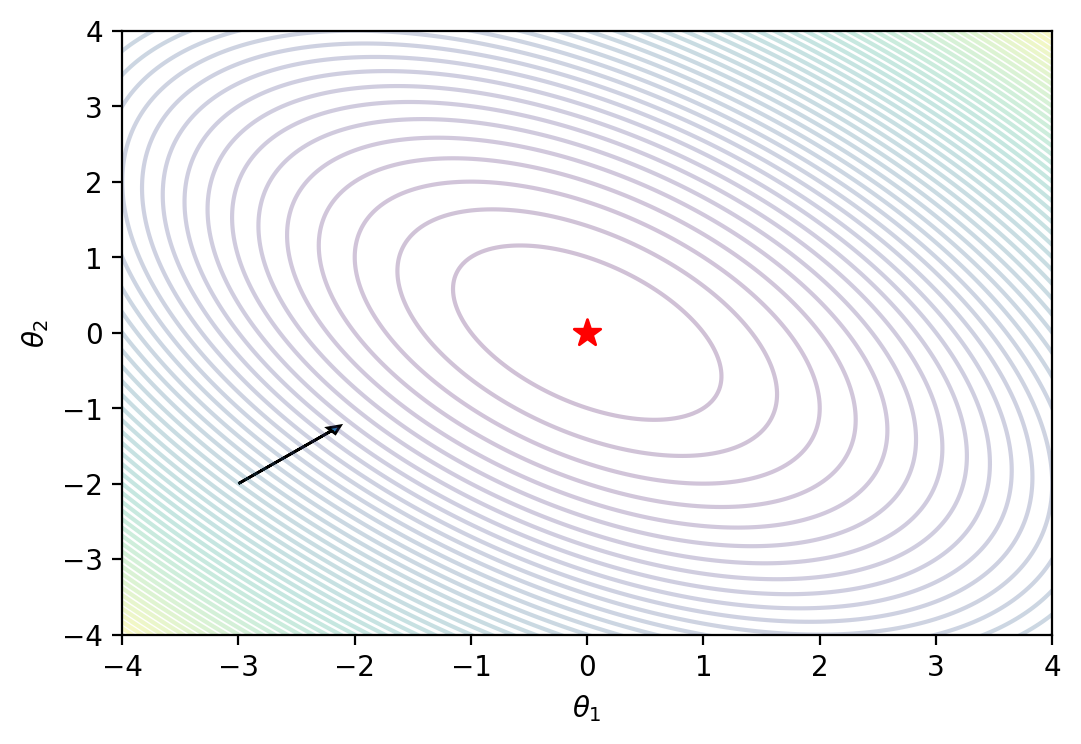}}
  \subfigure[Gradient Accumulation]{\label{fig:agn_intuition_b}\includegraphics[width=.45\textwidth]{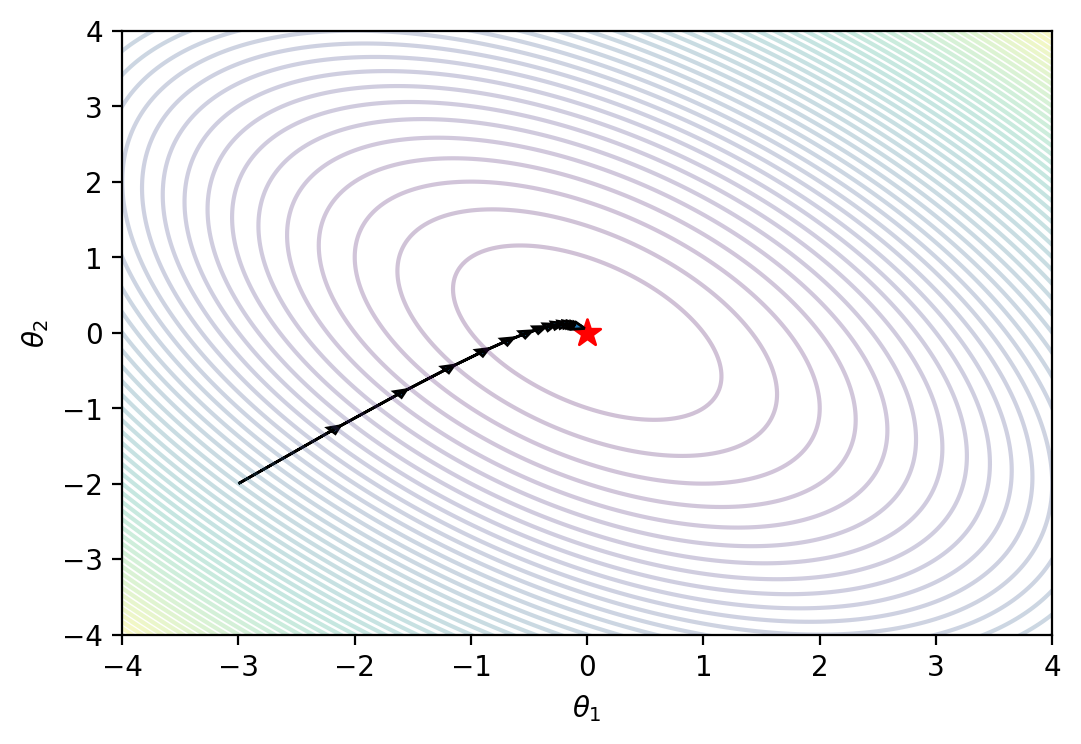}}
  \caption{This figure shows the difference between regular first-order gradients (a), and accumulated gradients (b). We observe that \emph{accumulated gradients are proportionally larger to the number of exploration steps}. However, the accumulated gradient does provide a better direction compared to first-order gradients.}
  \label{fig:agn_intuition}
\end{figure}

The main issue with \textsc{downpour}~\cite{dean2012large} is the requirement of constant communication with the parameter server after every gradient computation. Furthermore, as the number of parallel workers increases, \textsc{downpour} fails to converge due to the amount of \emph{implicit momentum}, as shown in Figure~\ref{fig:downpour_convergence}. To reduce the amount of communication with the parameter server, one could take ideas from \textsc{easgd}, and perform several iterations of local exploration before committing the gradients to the parameter server. However, in the case of algorithms like \textsc{downpour}, that is, where gradients are committed to the parameter server in an asynchronous fashion with no mechanisms in place to ensure convergence, more local exploration results in proportionally larger gradients and as a result, complicate the staleness and the implicit momentum problem even further. To intuitively show why this is an issue, let us consider Figure~\ref{fig:agn_intuition}. In a \textsc{downpour} setting, first-order gradients such as in Subfigure (a) are committed to the parameter server. However, when an algorithm allows for a certain amount of local exploration, the gradient that is committed to the parameter server is typically an \emph{accumulated gradient} as shown in Subfigure (b).\\

\begin{figure}
  \centering
  \subfigure[$\lambda = 1$]{ \includegraphics[width=.45\linewidth]{resources/images/downpour_10}}
  \subfigure[$\lambda = 20$]{ \includegraphics[width=.45\linewidth]{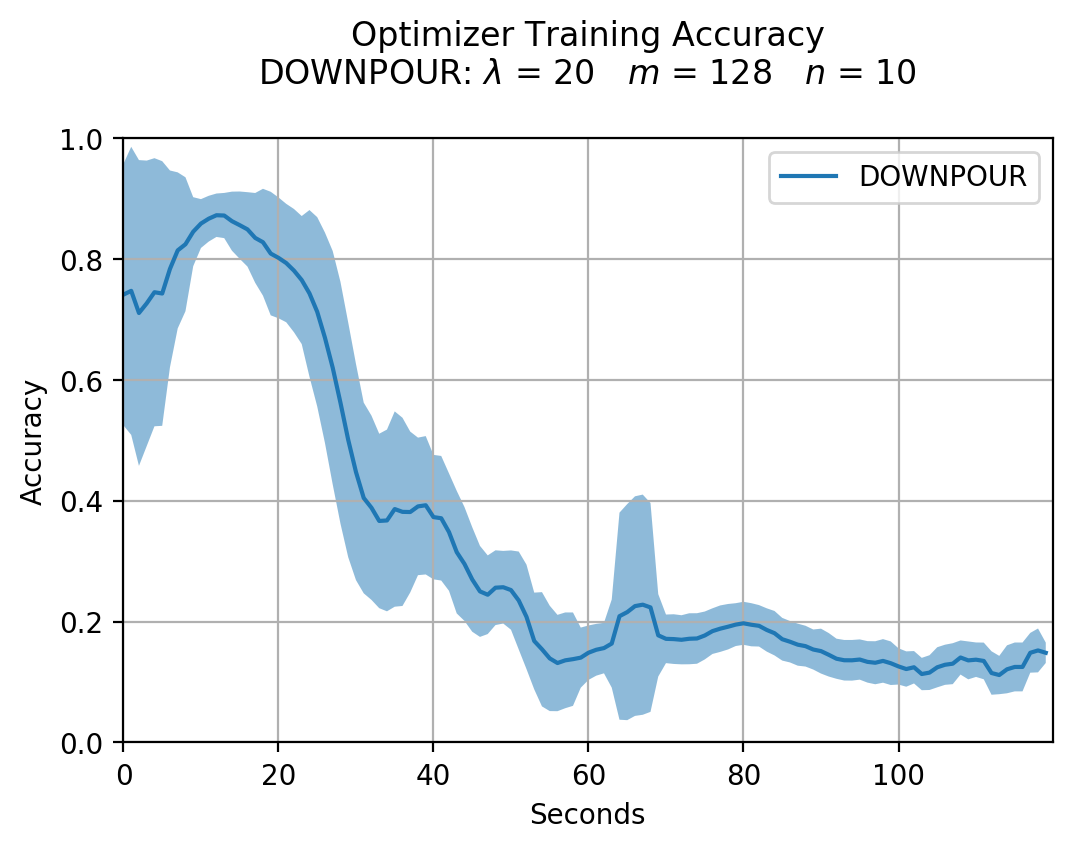}}
  \caption{Illustration of divergence due to gradient accumulation in \textsc{downpour}. In Figure~\ref{fig:downpour_convergence}, we say that for $n = 10$ \textsc{downpour} converged to a good solution. In order to reduce the training time, we decrease the communication frequency (increasing $\lambda$). However, due to the larger gradients that are committed to the parameter server, which increases the amount of implicit momentum, the central variable is not able to converge as before.}
  \label{fig:downpour_accumulated_divergence}
\end{figure}

Now, imagine two asynchronous environments where respectively no gradient accumulation takes place, and one where does. In the environment where no gradient accumulation is performed, as in regular \textsc{downpour}, first-order gradients are committed to the parameter server. However, we know that \textsc{downpour} diverges when the number of asynchronous workers is too high due to the amount of implicit momentum~\cite{implicitmomentum}. As a result, careful tuning is required when no adaptive methods are applied in order to guarantee convergence. Nevertheless, given the fact that \textsc{downpour} converges with $n = 10$ workers in Figure~\ref{fig:downpour_convergence}, and our knowledge about gradient accumulation, i.e., \emph{accumulated gradients that are committed are proportional to the number of exploration steps for every worker, and provide better directions to a minimum}, we would expect that for some amount of local exploration while using the same hyperparameterization (with the exception of local exploration steps $\lambda$) \textsc{downpour} would diverge again due to the magnitude of the accumulated gradients. This behaviour is illustrated in Figure~\ref{fig:downpour_accumulated_divergence}, and confirms our hypothesis.\\

To reduce the magnitude of the accumulated gradients, and thereby reducing the amount of implicit momentum, while at the same time preserving the better direction that has been provided due to the amount of local exploration, we propose to normalize (average) the accumulated gradient with the amount of local steps that have been performed by the workers ($\lambda$), shown in Equation~\ref{eq:accumulated_gradient_normalization}\footnote{Note if $\lambda = 1$, \textsc{agn} generalizes to \textsc{downpour}.}. We call this technique of normalizing the accumulated gradient \emph{Accumulated Gradient Normalization} or \textsc{agn}. An initial critique of this technique would be that by normalizing the accumulated gradient, \textsc{agn} would in effect be undoing the work that has been done by a single worker. This seems at first a valid criticism, however, one needs to take into account that \textsc{agn} is actually using the worker exploration steps to compute a better gradient based on first-order gradients.

\begin{figure}
  \centering
  \includegraphics[width=.45\textwidth]{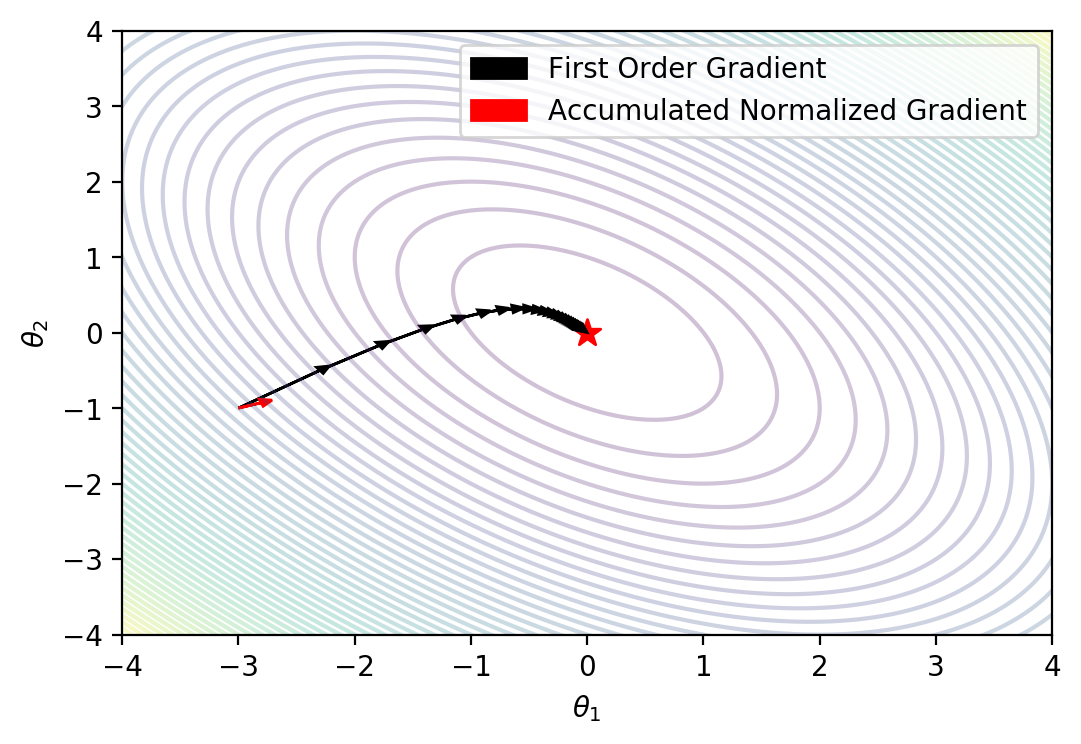}
  \caption{After pulling the most recent parameterization of the central variable from the parameter server, the worker starts accumulating $\lambda$ first order gradients, and applies those gradients locally to explore the surrounding error space. Finally, after $\lambda$ exploration steps have been performed, the accumulated is normalized w.r.t. $\lambda$ and send to the parameter server.}
    \label{fig:agn_example}
\end{figure}

\section{Method}
\label{sec:method}

The \textsc{agn} update rule for a worker and the parameter are described in Equation~\ref{eq:accumulated_gradient_normalization} and Equation~\ref{eq:agn_ps_update} respectively. Since \textsc{agn} is using local steps ($\lambda$) to compute a better gradient based on a \emph{sequence} of first-order gradients, it can also be used under communication constraints like \textsc{easgd} since less communication with the parameter server is required due to computation of a local sequence of first-order gradients. Figure~\ref{fig:agn_example} shows how an \textsc{agn} gradient is obtained and computed using Equation~\ref{eq:accumulated_gradient_normalization} by following Algorithm~\ref{algo:agn}. In this setting $\eta_t$ denotes the learning rate of a worker at time $t$, and $m$ denotes the size of the mini-batch which is identical across all workers.

\begin{equation}
  \label{eq:accumulated_gradient_normalization}
  \Delta\theta^k = -\frac{1}{\lambda}\sum_{i = 0}^\lambda \eta_t \frac{1}{m}\sum_{j = 0}^{m - 1} \nabla_\theta \mathcal{L}(\theta_i;x_{ij};y_{ij})
\end{equation}

\begin{equation}
  \label{eq:agn_ps_update}
  \tilde{\theta}_{t + 1} = \tilde{\theta}_t + \Delta\theta^k
\end{equation}

\begin{algorithm}
  \caption{Worker procedure of \textsc{agn}.}
  \label{algo:agn}
  \begin{algorithmic}[1]
    \Procedure{AGNWorker}{$k$}
    \State $\theta^k_0 \gets \tilde{\theta} \gets \Call{Pull}$
    \State $t \gets 0$
    \State \textbf{while} {$\textbf{not}$ converged} \textbf{do}
    \State ~~$i \gets 0$
    \State ~~$a \gets 0$
    \State ~~\textbf{while} $i < \lambda$ \textbf{do}
    \State ~~~~$\textbf{x},~\textbf{y} \gets \Call{FetchNextMiniBatch()}{}$
    \State ~~~~$g \gets -\eta_t \odot \nabla_\theta \mathcal{L}(\theta^k_t;\textbf{x};\textbf{y})$ \Comment{Gradient from, e.g., \textsc{Adam}~\cite{kingma2014adam}}
    \State ~~~~$a \gets a + g$
    \State ~~~~$\theta^k_{t + 1} = \theta^k_t + g$
    \State ~~~~$i \gets i + 1$
    \State ~~~~$t \gets t + 1$
    \State ~~\textbf{end}
    \State ~~$a \gets \frac{a}{\lambda}$ \Comment{Normalization step.}
    \State ~~$\Call{Commit}{a}$
    \State ~~$\theta^k_{t} \gets \Call{Pull}$
    \State \textbf{end}
    \EndProcedure
  \end{algorithmic}
\end{algorithm}

An interesting thought-experiment would be to explore what would happen if the workers would communicate with the parameter server after a very large number of steps, that is, when $\lambda$ approaches $\infty$. How would the normalized accumulated gradients look like in such a situation, described by Equation~\ref{eq:agn_thought_experiment}?

\begin{equation}
  \label{eq:agn_thought_experiment}
  \lim_{\lambda \to \infty} -\frac{\sum_{i = 0}^\lambda \eta_t \frac{1}{m}\sum_{j = 0}^{m - 1} \nabla_\theta \mathcal{L}(\theta_i;x_{ij};y_{ij})}{\lambda}
\end{equation}

In order to completely understand how the worker deltas would look like after $\lambda = \infty$ steps, one first needs to understand the individual components of Equation~\ref{eq:agn_thought_experiment}. The most inner component, $\eta_t \frac{1}{m}\sum_{j = 0}^{m - 1} \nabla_\theta \mathcal{L}(\theta_i;x_{ij};y_{ij})$, is just the computation of a mini-batch using $m - 1$ samples, where index $i$ denotes the current step in the gradient accumulation. Please note that a mini-batch can differ for different values of $i$ as training samples are randomly retrieved from the dataset. After computing the gradient based on the mini-batch, the local model will be updated as $\theta_{i + 1} = \theta_i - \eta_t\frac{1}{m}\sum_{j = 0}^{m - 1} \nabla_\theta \mathcal{L}(\theta_i;x_{ij};y_{ij})$. This process goes on for $\lambda$ steps, while at the end, the accumulated is normalized with respect to $\lambda$.\\

\begin{figure}
  \centering
  \subfigure[$\lambda = 15$]{\includegraphics[width=.45\linewidth]{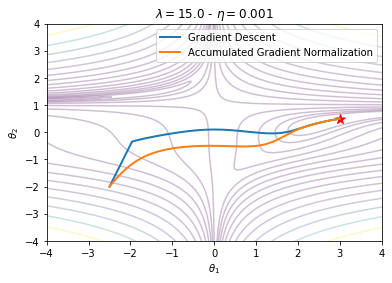}}
  \subfigure[$\lambda = \infty$]{\includegraphics[width=.45\linewidth]{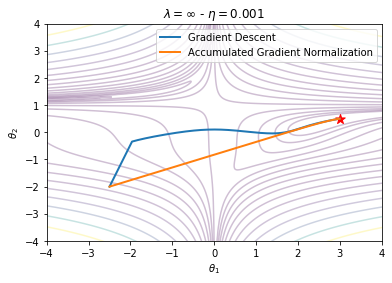}}
  \caption{\textsc{agn} for different values of $\lambda$. This small experiment shows that when $\lambda = \infty$, the path the central variable traverses is equal to a straight line towards the minima.}
  \label{fig:agn_straight_line}
\end{figure}

Let us assume we have a smooth convex error space, or a smooth non-convex error space with at least a single minima. Due to the existence of a minima in both cases, first order gradients will eventually converge to, or in the neighbourhood of said minima. Furthermore, we  make the assumption that the hyperparameterization during the training procedure will not change. For instance, no learning rate decay after $x$ number of steps. Under these assumptions, it is trivial to realize that applying gradient descent for $\infty$ steps will cause the parameterization to converge in a minima. Of course, given that the hyperparameterization, and the data allow for convergence to occur. As a result, the term $\sum_{i = 0}^\lambda \eta_t \frac{1}{m}\sum_{j = 0}^{m - 1} \nabla_\theta \mathcal{L}(\theta_i;x_{ij};y_{ij})$ is finite, even after applying $\infty$ steps of mini-batch updates. To simplify our problem, let us denote $\vec{c}$ as the \emph{finite} result of the top term in Equation~\ref{eq:agn_thought_experiment} for $\lambda = \infty$. Furthermore, since $\vec{c}$ is finite, the equation can be treated as an instance of $\frac{1}{\infty}$, which approaches 0. This implies that for a very large $\lambda$, the normalized accumulated gradients will basically be $\vec{0}$. However, what is interesting is that the normalized accumulated gradients directly point towards a minima due to the large amount of exploration steps that have been performed. Subsequently, one can view a normalized accumulated gradient when $\lambda$ approaches $\infty$ as a point, but with a direction. Therefore, when convergence is obtained, the path the central variable traversed is a straight line towards the minima, as shown in Figure~\ref{fig:agn_straight_line}.\\

The thought experiments described above help us in several ways if we make several additional assumptions. The first assumes that normalized accumulated gradients with $\lambda = \infty$ can be computed immediately, that is, without a delay. This is of course an unrealistic assumption. However, one needs to consider realistic communication constraints. Given a certain network throughput, what is the amount of local communication that needs to be performed in order for a parameter commit to be ``worth it''? As mentioned above, $\lambda = \infty$ is not a very good solution since the normalized accumulated gradient will converge to $\vec{0}$ in the limit. Nevertheless, if the normalized accumulated gradient could be computed immediately, as we assumed, the central variable would traverse the shortest path to a minima, in contrast to first order gradients. Of course, this is not a realistic assumption. Furthermore, this issue is quite similar to \emph{stochastic gradient descent} vs. \emph{mini-batch gradient descent}, since in \textsc{agn} we also have to make the decision between more frequent parameter updates, and more ``local'' iterations to compute a better gradient, where better in the case of mini-batch gradient descent means less-noisy.\\

In most settings, the size of a mini-batch is determined empirically, and is dependent on the noise of the gradients. Furthermore, when using mini-batch gradient descent, a trade-off is made between more frequent parameter updates, i.e., a smaller mini-batch, or more robust and consistent gradients by increasing the size of a mini-batch which results in a more accurate approximation of the first order curvature. This is similar to our situation. Yet, in mini-batch gradient descent you are basically trying to estimate a hyperparameter based on several unknowns, i.e., convergence based on error space and noise of individual gradients. However, \textsc{agn} is balancing the amount of local computation to produce a better gradient, with the throughput of the network, which is a known variable. For instance, imagine a hypothetical communication infrastructure which is able to apply the commits of the workers directly into the workers with no delay. In this situation, one could apply \textsc{downpour}. However, remember from Figure~\ref{fig:downpour_accumulated_divergence} that \textsc{downpour} does not handle an increased amount of asynchronous parallelism ($n = 20$). As a result, even in an ideal situation \textsc{downpour} will not be able to converge due to the amount of implicit momentum.\\

Nevertheless, the situation in \textsc{agn} is different as will become apparent in Section~\ref{sec:experiments}. Contrary to \textsc{downpour}, \textsc{agn} does not commit first order gradients to the parameter server, but rather a normalized sequence of first order gradients which result in better directions towards a minima, as discussed above. Therefore, \textsc{agn} worker deltas will point more or less in more optimal direction and thereby reducing the negative effects of implicit momentum in first order gradients.

\section{Experimental Validation}
\label{sec:experiments}

This Section evaluates \textsc{agn} against different distributed optimization algorithms. MNIST~\cite{mnist} is used as a benchmark dataset, and all optimizers the same model with identical parameterization of the weights. Furthermore, we will set the mini-batch size to $m = 128$ in all optimizers, and use \emph{40 epochs} worth of training data that will be equally distributed over all $n$ workers. Our computing infrastructure consists a relatively small cluster of \emph{15 nodes} with a \emph{10Gbps interconnect}, most of them in the same rack, each having 2 Intel\textsuperscript{\textregistered} Xeon\textsuperscript{\textregistered} CPU E5-2650 v2 @ 2.60GHz CPU's, where every CPU has 8 cores and 2 threads. No GPU's are used during training, and no learning rate decay is applied. Our optimizer and experiments are implemented and executed using \emph{dist-keras}\footnote{\url{github.com/cerndb/dist-keras}}, which are available in the package.\\

\begin{figure}
  \centering
  \subfigure[\textsc{dynsgd} $\lambda = 10$]{\includegraphics[width=.45\textwidth]{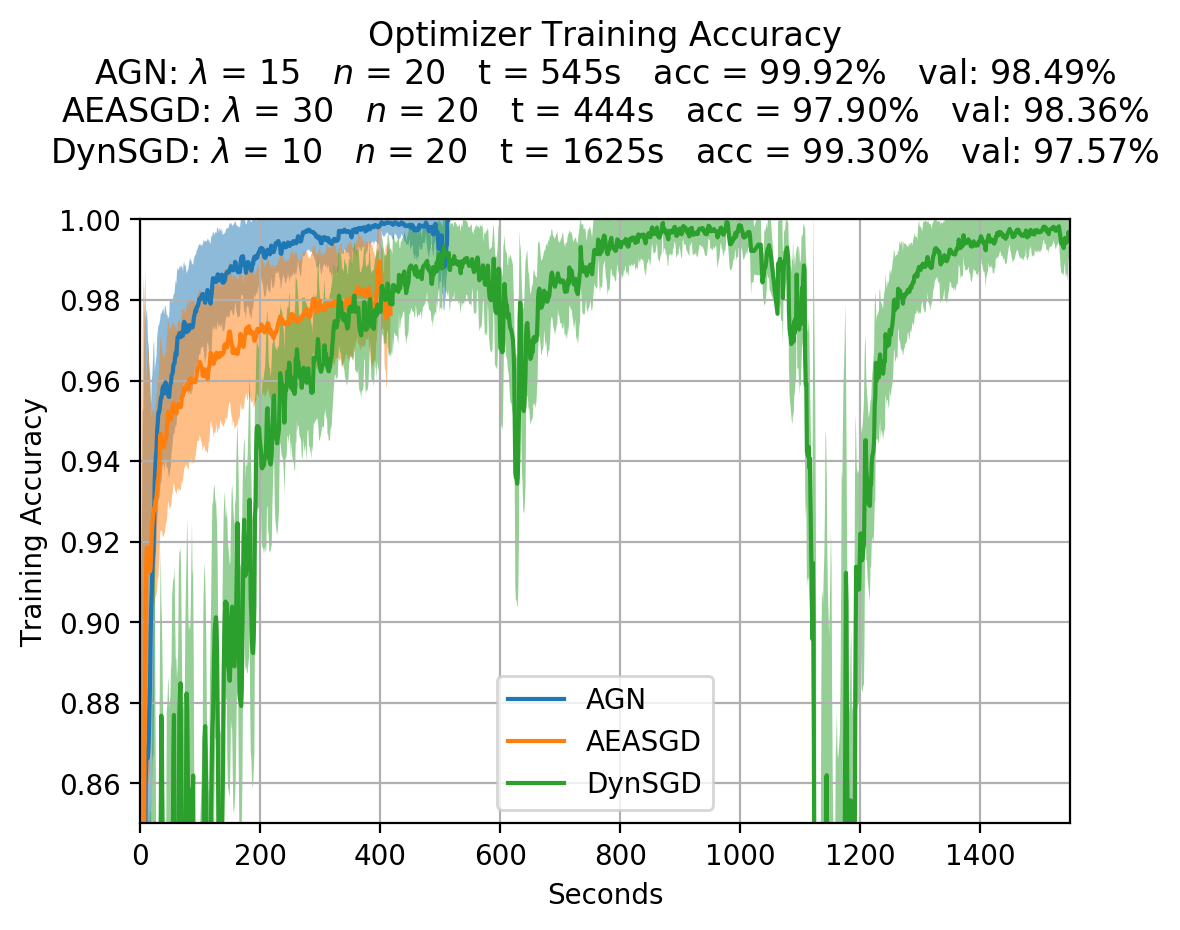}}
    \subfigure[\textsc{dynsgd} $\lambda = 3$]{\includegraphics[width=.45\textwidth]{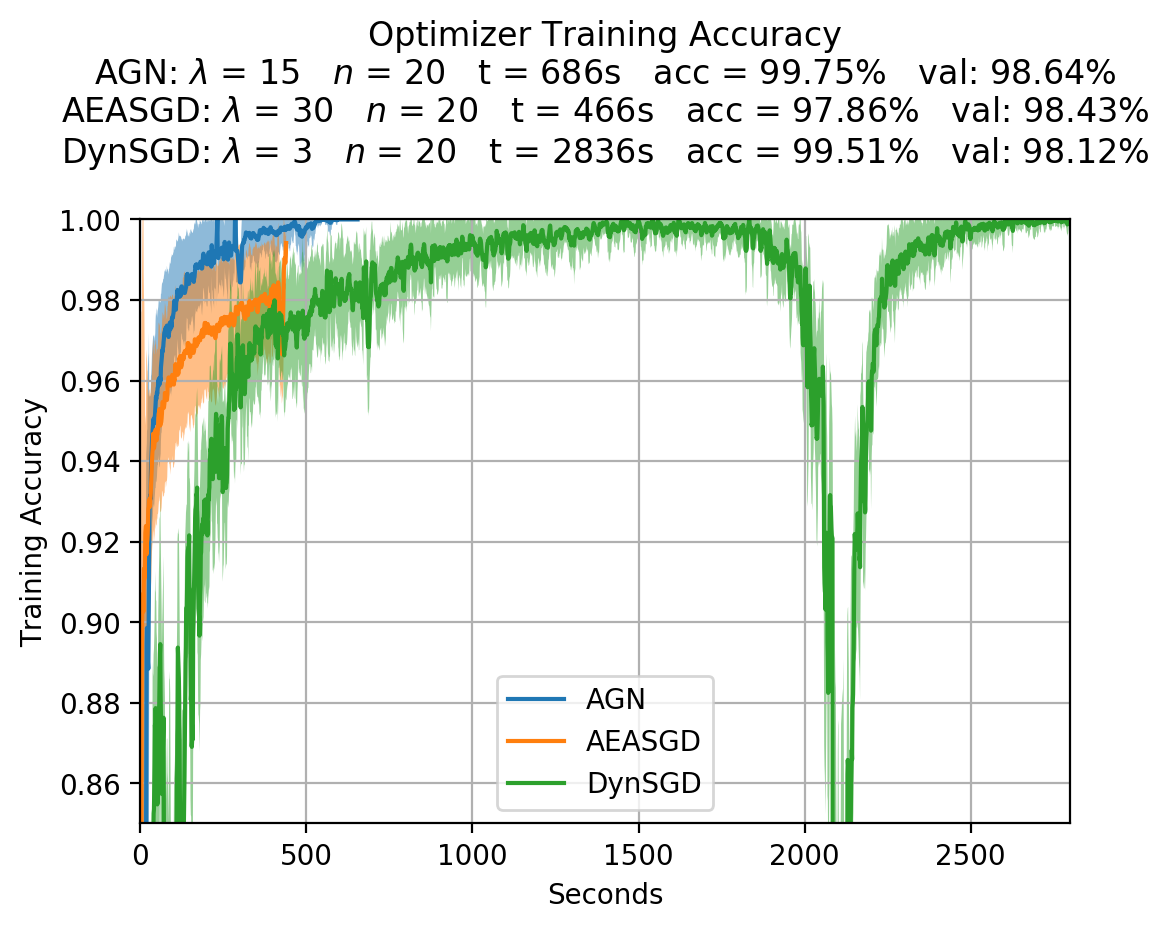}}
  \caption{In this experiment we train all optimizers on 40 epochs worth of data with a mini-batch of $m = 128$. We observe that \textsc{agn} significantly outperforms all other optimizers. Furthermore, due to staleness-handling method of \textsc{dynsgd}, the optimizer is not able to handle accumulated gradients which results in non-stale accumulated gradients being incorporated directly into the central variable with the disadvantage that other workers are even more stale in terms of parameter distance. Which is the root cause of this divergent behaviour. In Subfigure (b) we reduce the amount of local exploration steps, which in turn reduces the length of the accumulated gradient. Therefore causing other workers to be less stale, and consequently reducing the divergent effects we observed in Subfigure (a).}
  \label{fig:agn_experiment_1}
\end{figure}

Our initial experiment, shown in Figure~\ref{fig:agn_experiment_1}, shows the training accuracy of \textsc{agn}, \textsc{aeasgd}, and \textsc{dynsgd} over time. In this experiment, we use a near-optimal hyperparameterization for all optimizers to ensure convergence. Looking at Figure~\ref{fig:agn_experiment_1}, we observe an increase in training performance for \textsc{agn}, both in training accuracy, and in training time when compared to current state-of-the-art algorithms such as \textsc{aeasgd} and \textsc{dynsgd}. Furthermore, \textsc{dynsgd} scales the gradients down with respect to staleness $\tau$, which in effect is $(n - 1)^{-1}$ since $\textbf{E}[\tau] = n - 1$. As a result, \textsc{dynsgd} does not handle the parameter staleness problem appropriately since it does not take the distance between the parameterizations into account. This is validated in Figure~\ref{fig:agn_experiment_1} because of the observed divergent behaviour of the optimizer. Furthermore, due to the relatively high communication frequency ($\lambda = 10$, $\lambda = 3$), \textsc{dynsgd} will take longer to process all data since more communication with the parameter server is required.\\

\begin{figure}
  \centering
  \includegraphics[width=.5\textwidth]{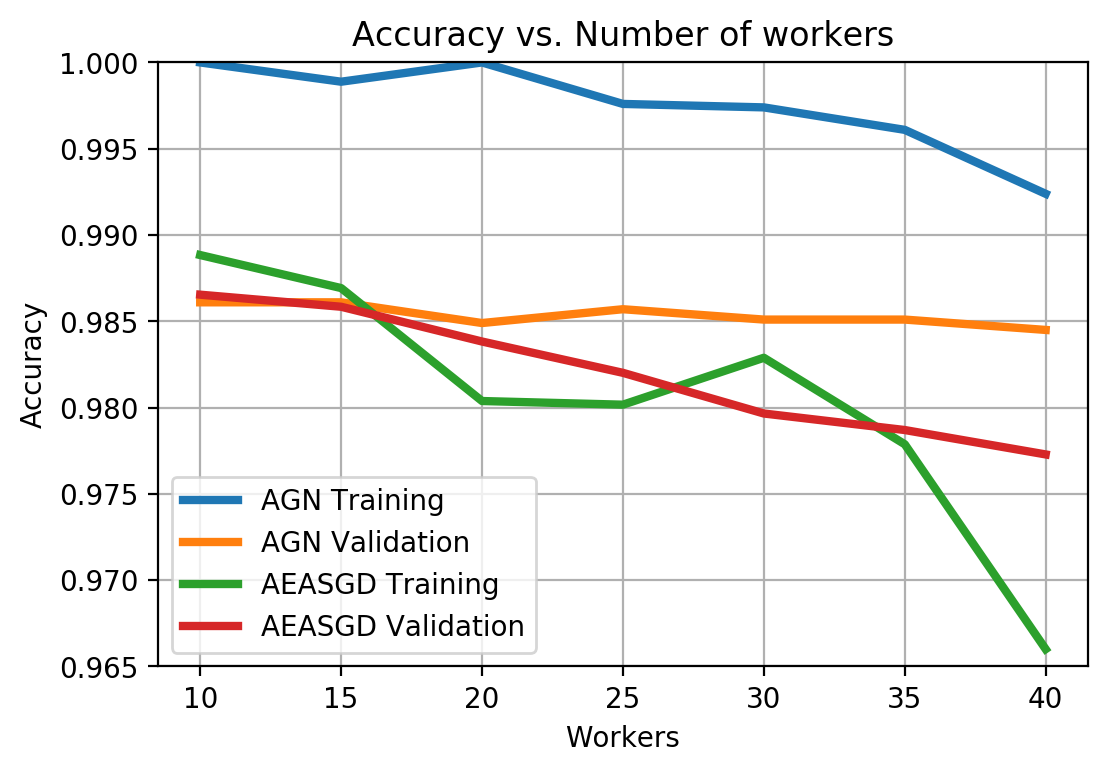}
  \caption{Decline of the training accuracy of both \textsc{agn} and \textsc{aeasgd} as the number of asynchronous workers increases. From these experiments, we observe that \textsc{agn} is more robust to an increased amount of asynchrony as the training accuracy only starts to decrease from $n = 25$ workers, while the validation accuracy remains stable even with 40 workers.}
  \label{fig:agn_aeasgd_workers_performance_decline}
\end{figure}

\begin{figure}
  \centering
  \subfigure[\textsc{agn} varying $\lambda$]{\includegraphics[width=.31\linewidth]{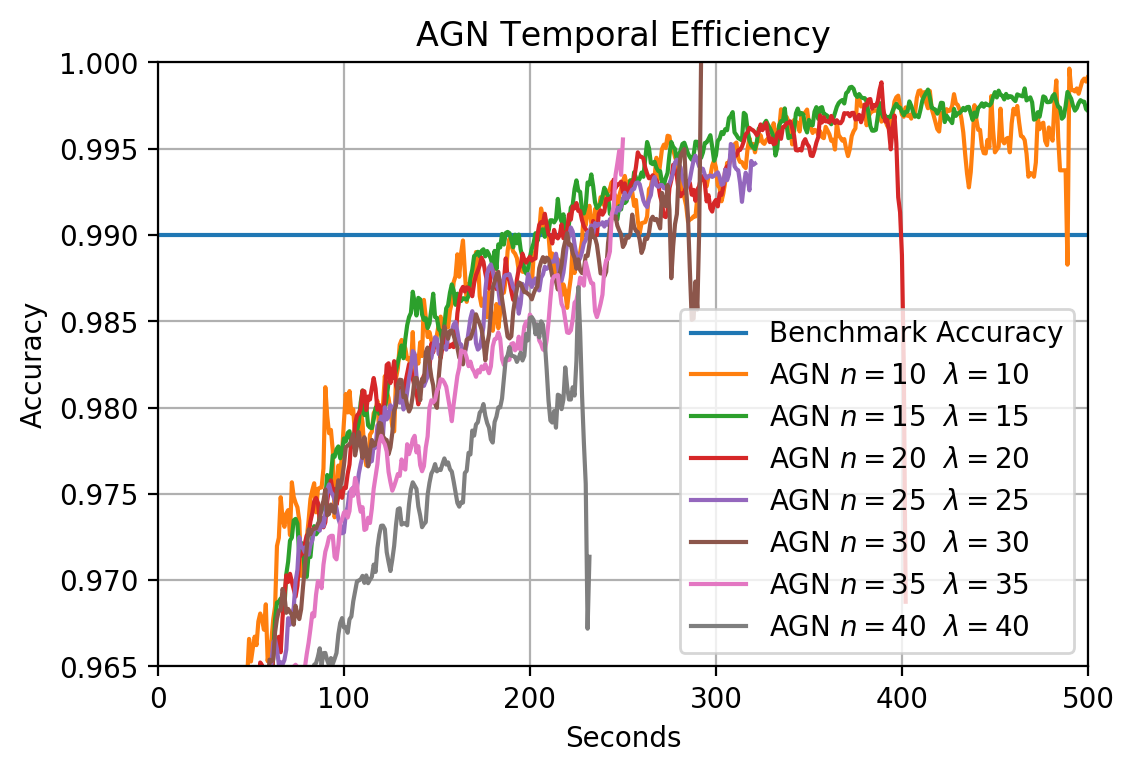}}
  \subfigure[\textsc{agn} static $\lambda$]{\includegraphics[width=.31\linewidth]{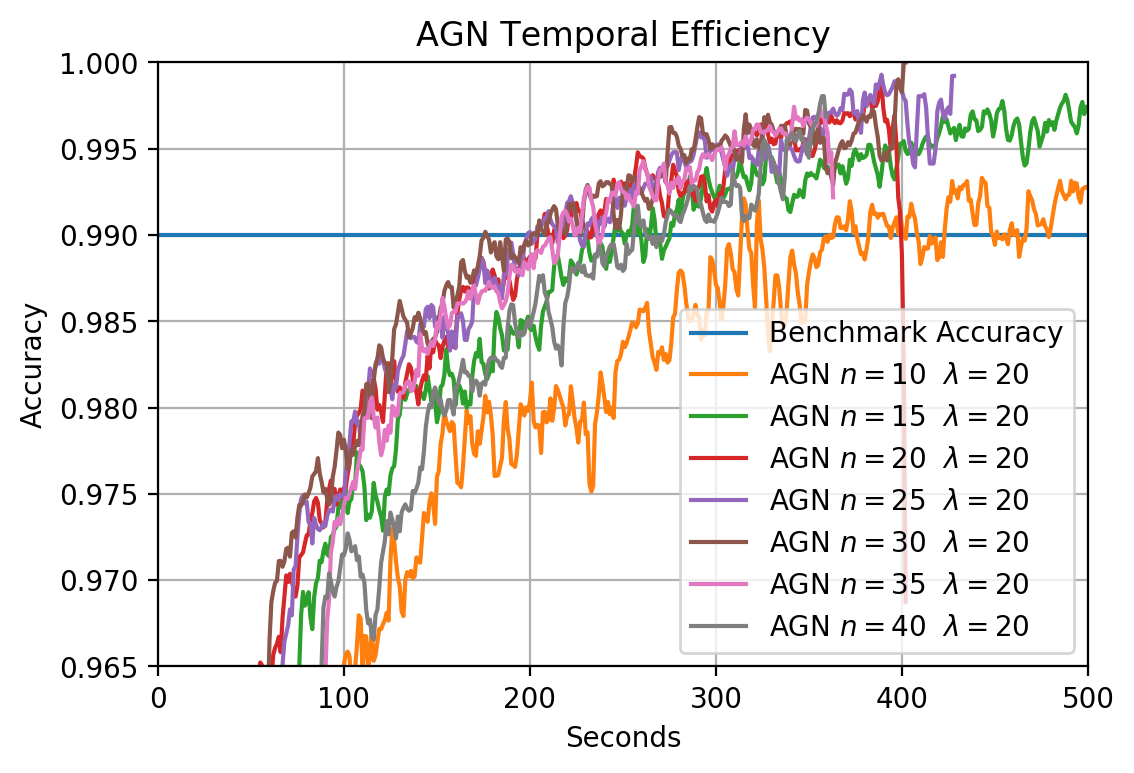}}
  \subfigure[\textsc{aeasgd} varying $\lambda$]{\includegraphics[width=.31\linewidth]{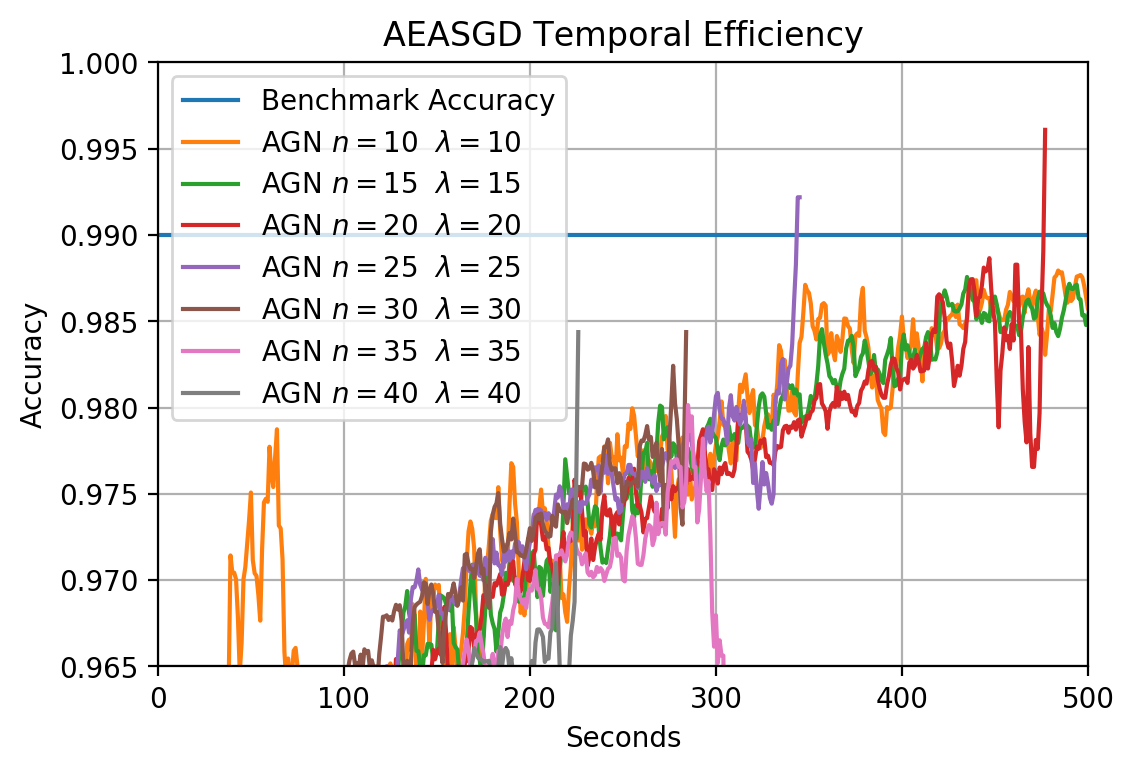}}
  \caption{Training accuracy plots for different configurations of the distributed hyperparameters. In the case of a varying $\lambda$ with respect to the number of workers (to minimize the noise of the commits), we observe that optimizers with a higher communication frequency (small $\lambda$), are actually benefiting from the more frequent updates with the parameter server. However, as the number of asynchronous workers grows, a low communication frequency increases the noise in the commits due to parameter staleness. Furthermore, if the $lambda$ is too large, less frequent parameter server updates occur, which results in a slower convergence rate since more time is spent locally. As a result, a balance is required similar to determining the size of a mini-batch.}
  \label{fig:agn_temporal_efficiency}
\end{figure}

Contrary to \textsc{aeasgd}, \textsc{agn} is able to cope more effectively with an increased amount of parallelism, as its training accuracy only starts to decline from $n = 25$ asynchronous workers, while the validation accuracy is barely fluctuating, as shown in Figure~\ref{fig:agn_aeasgd_workers_performance_decline}. An obvious follow-up question to this result would be to question the fact whether increasing the amount of workers really improves the temporal efficiency optimizer, i.e., the amount of time it takes to reach a certain training accuracy. In fact, it does reduce the training time to reach a certain training accuracy, as shown in Figure~\ref{fig:agn_temporal_efficiency}. However, several factors have to be taken into account.\\

\begin{figure}
  \centering
  \subfigure[$n = 10$]{ \includegraphics[width=.3\linewidth]{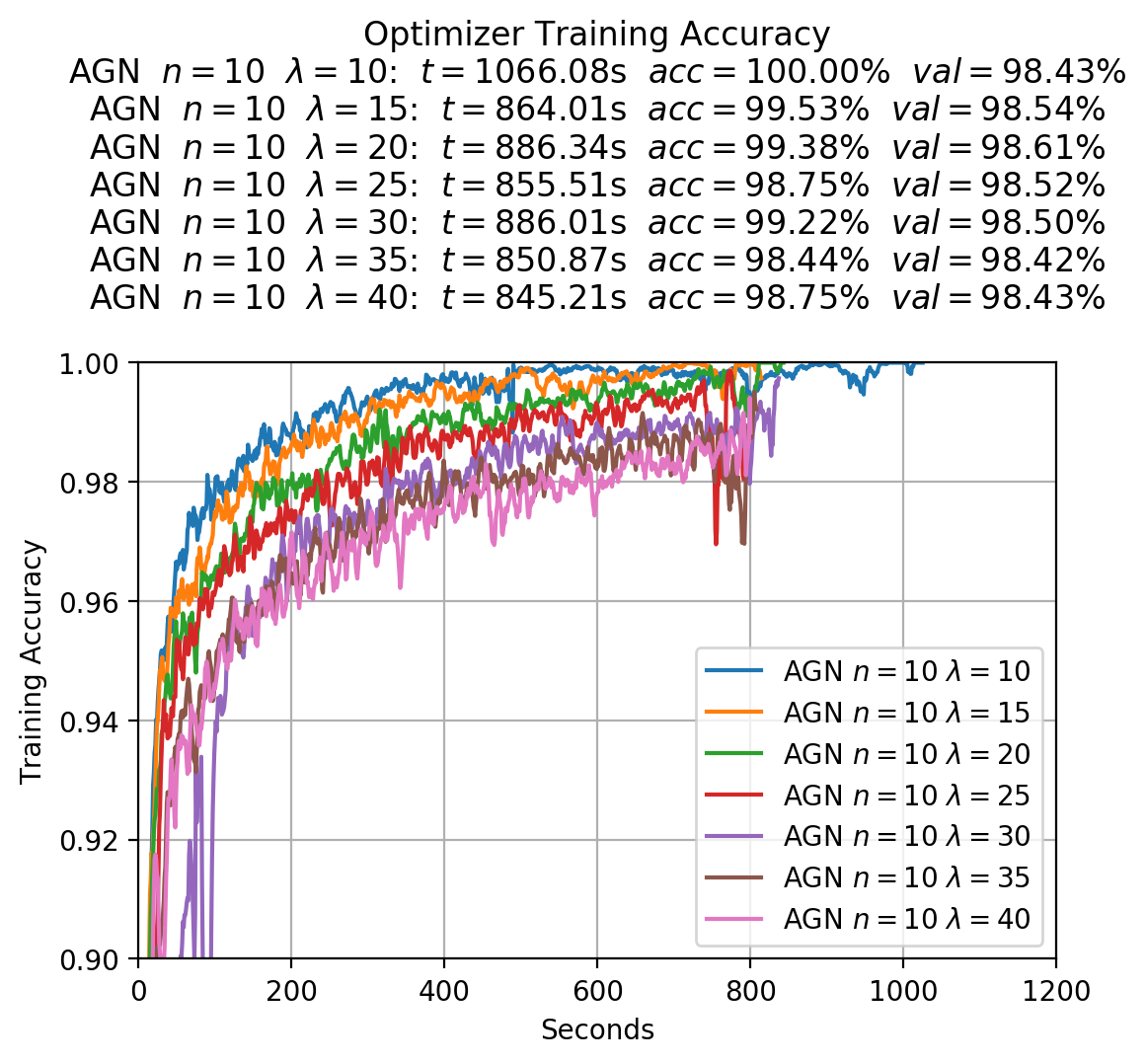}}
  \subfigure[$n = 15$]{ \includegraphics[width=.3\linewidth]{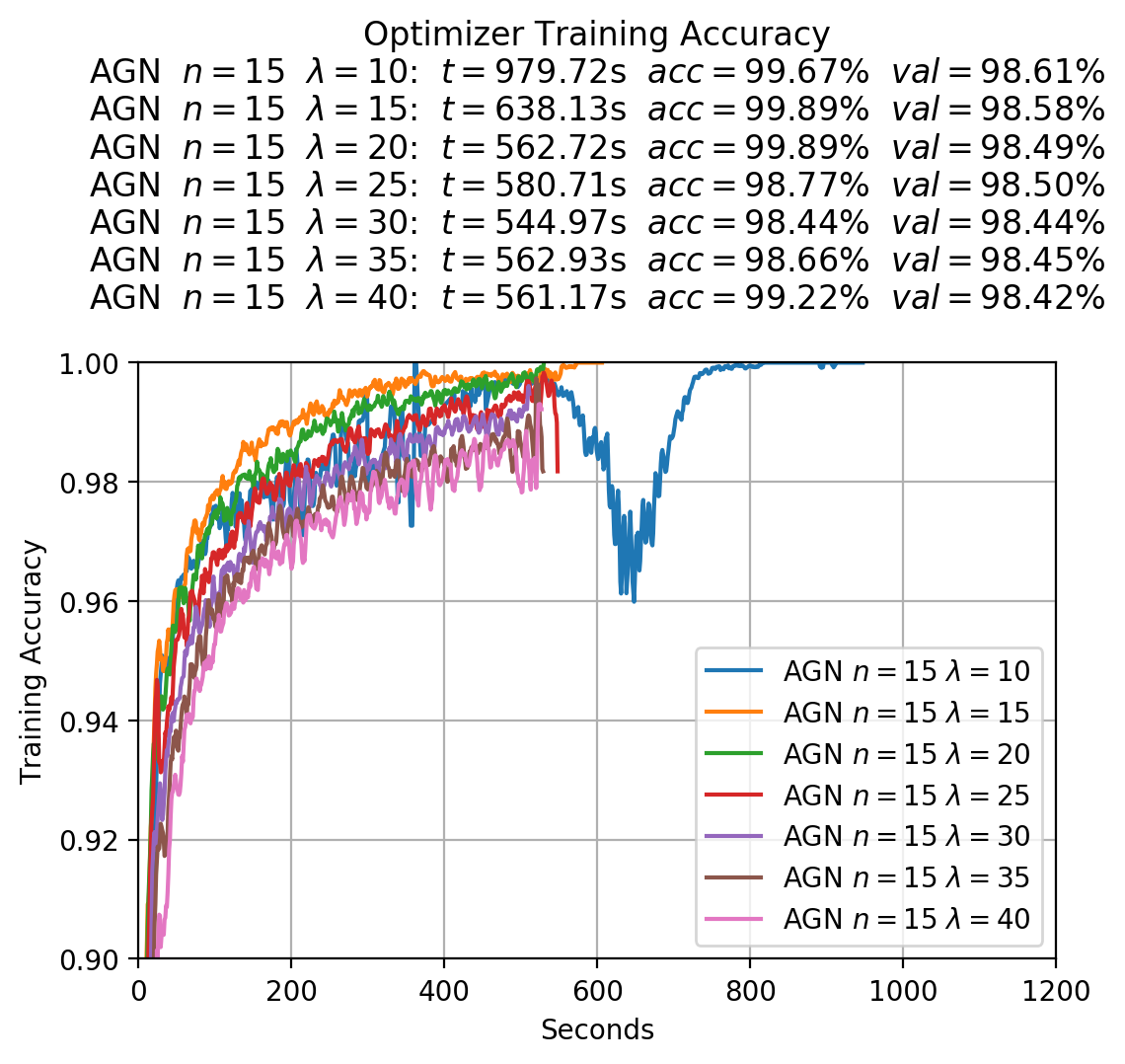}}
  \subfigure[$n = 20$]{ \includegraphics[width=.3\linewidth]{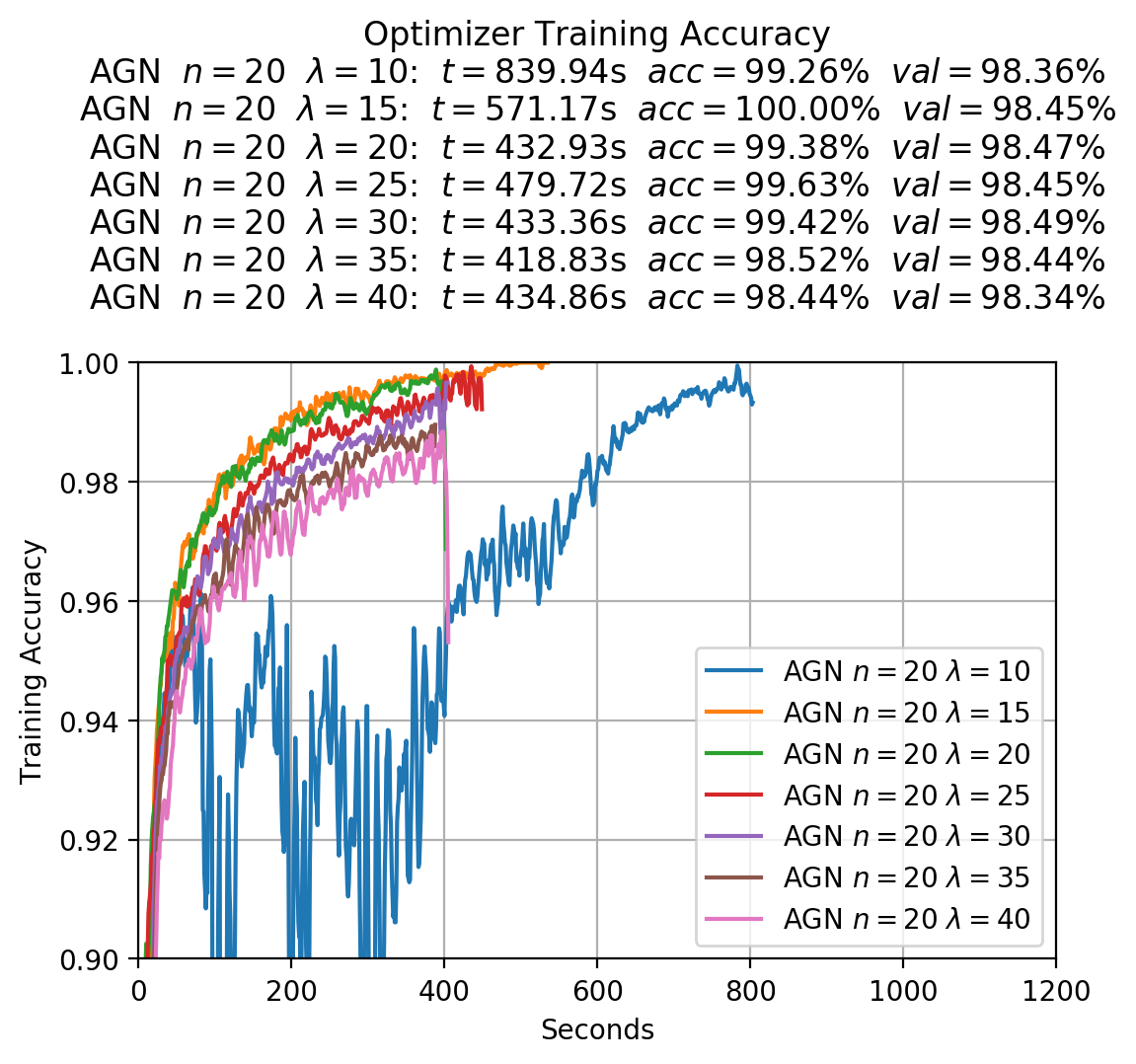}}
  \subfigure[$n = 25$]{ \includegraphics[width=.3\linewidth]{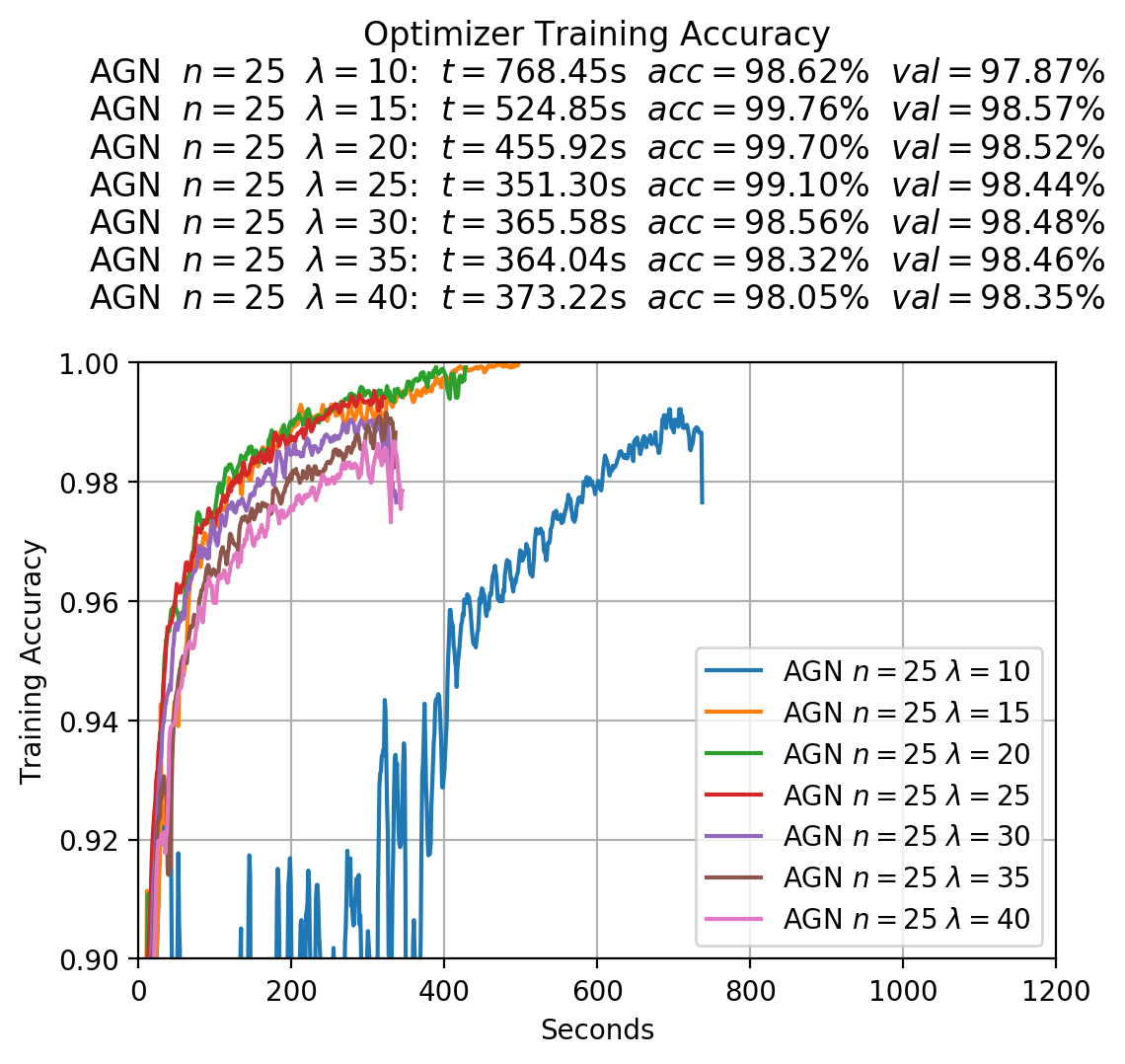}}
  \subfigure[$n = 30$]{ \includegraphics[width=.3\linewidth]{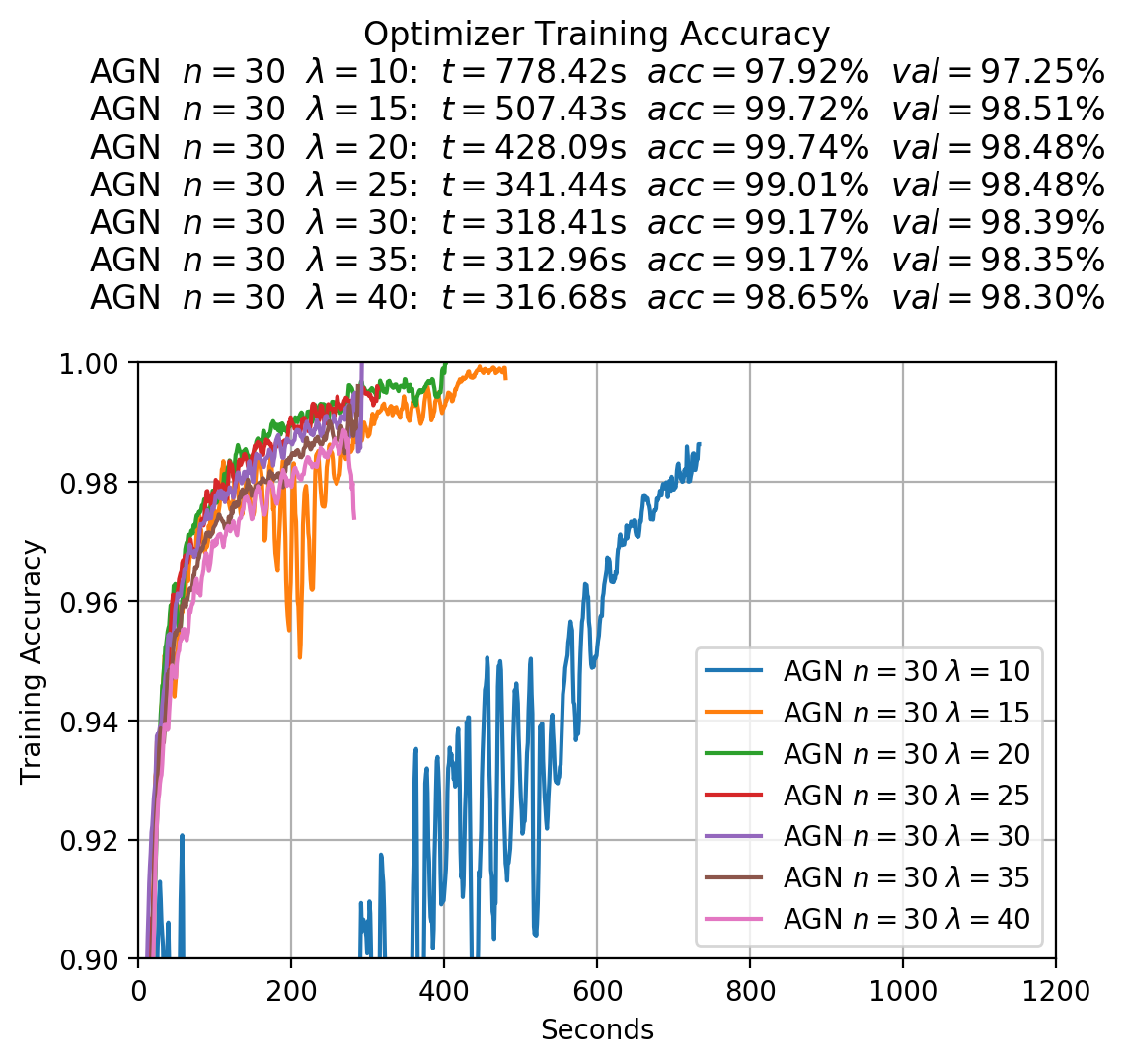}}
  \subfigure[$n = 40$]{ \includegraphics[width=.3\linewidth]{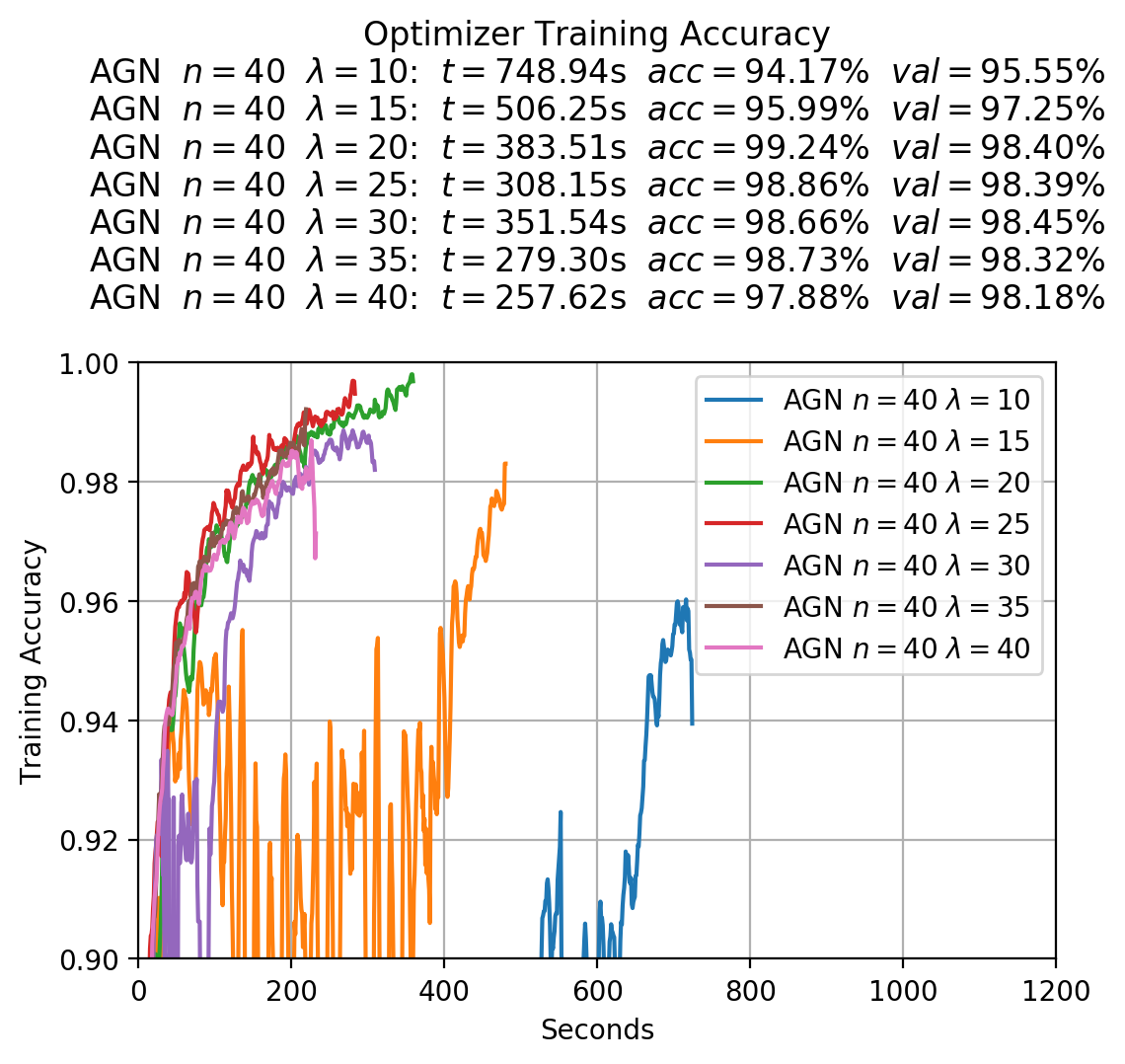}}
  \caption{This Figure shows several experiments were we fixed the number of workers, but vary the communication frequency. From this we observe that \textsc{agn} performs well when a relatively equal high communication frequency is used with respect to the number of workers. Furthermore, increasing the amount of workers, and maintaining a high communication frequency deteriorates the performance of the central variable as well. As a result, a balance between the communication frequency, and the number of asynchronous workers is required.}
  \label{fig:agn_experiments_workers}
\end{figure}

The first being an increased amount of staleness that is inserted into the system as the number of asynchronous workers increase. This effect is difficult to mitigate. Previous approaches~\cite{jiang2017heterogeneity} propose to scale gradient commits down proportionally to the number of stale steps. However, as previously shown, this is not an effective solution since accumulating gradients locally, is in effect making the gradients larger, and as a result, committing accumulated gradients increases the \emph{distance} between the central variable and other workers. The second and final issue is the balance between updating the central variable with a certain frequency, and the amount of local work to effectively reduce the training time due to high communication costs. In effect, this resembles the situation usually one has when selecting a mini-batch size $m$, i.e., do we allow for more frequent parameter updates (small $m$), or do we compute a less noisy first order gradient by increasing $m$, thereby reducing the frequency of parameter updates and the convergence of a model? In Figure~\ref{fig:agn_experiments_workers}, we evaluate varying values of $\lambda$ for a specific number of asynchronous workers $n$ to show that similar manual tuning is required. In all cases, we observe configurations with $\lambda = 40$ usually have the slowest convergence rate with respect to other configurations with higher communication frequencies.\\

Nevertheless, what is really interesting, is why configurations with low communication frequencies actually do converge, in contrast to configurations with high communication frequencies (with respect to the number of workers). Since our definition of staleness relates to the distance between the \emph{current} parameterization of a worker, and the \emph{current} parameterization of the central variable. One can imagine that increasing the number of asynchronous workers, effectively increases the \emph{distance} between the parameterizations of the workers and the \emph{current} central variable due to the queuing model discussed before, i.e., worker deltas are incorporated in the central variable in a queuing fashion. Yet, parameter staleness still does not explain why configurations with low communication frequencies converge, as opposed to configurations with higher communication frequencies. The question begs, is convergence guaranteed due to the amount of local exploration, thus providing the parameter server with a better ``direction'', as shown in Figure~\ref{fig:agn_example}. Or, due to limit condition described above, which eventually scales the worker deltas down to 0 as the communication frequency decreases ($\lambda$ increases)? This is a rather difficult question to answer, since there might be a synergy since large gradients are usually considered a bad thing. However, the normalization step takes the average gradient of all accumulated gradients. As a result, the limit condition does not apply here. Summarized, the stability property of \textsc{agn} arises from the fact that \emph{implicit momentum} fluctuations will mostly be in line with a minima due to the better directions \textsc{agn} provides.\\

To compare \textsc{agn} against \textsc{aeasgd} (since this optimizer shows almost no divergent behaviour for a wide range of hyperparameters), we introduce \emph{temporal efficiency} in terms of the surface described by a training metric. This means that for some optimizer $a$, we have a function $f_a(t)$ which describes the performance of a model at time $t$, e.g., $f_a(t)$ describes the training accuracy of the model at time $t$. If we integrate over $t$, we obtain a surface representing the performance of a model over time. If we would do this for an other optimizer $b$, and divide the surface of optimizer $a$ by the performance surface of optimizer $b$, we get a ratio which describes how optimizer $a$ is performing compared to optimizer $b$. If this ratio is larger then 1, it means that optimizer $a$ is outperforming optimizer $b$, else, it is the other way around (unless the surfaces are equal of course). However, in order to compute a \emph{fair} surface area, we have to limit the computation to the \emph{minimal shared training time} $m$. This is done to prevent that optimizers with a longer training time have a significant advantage, since they have more time to produce a better model. To summarize, we define the temporal efficiency $\mathcal{E}$ of two optimizers $a$ and $b$ as the ratio of their performance surface, as stated in Equation~\ref{eq:temporal_efficiency}. Using \emph{temporal efficiency}, we can make a more qualitative judgment which optimizer is performing better in different scenarios \emph{since it also incorporates the stability of the optimizer}.

\begin{equation}
  \label{eq:temporal_efficiency}
  \mathcal{E}(a,b) = \ddfrac{\int_0^m f_a(t) \,dt}{\int_0^m f_b(t) \,dt}
\end{equation}

Finally, we apply \emph{temporal efficiency} to compare \textsc{agn} against \textsc{aeasgd} and summarize the results in Table~\ref{table:agn_experiments_summary}. We make the observation that increasing the amount of asynchrony results in a deterioration of the training accuracy (which is expected since more staleness, and thereby, implicit momentum is induced). However, the rather unexpected property is that increasing the amount of asynchronous workers results in an early \emph{flattening} of the training accuracy in \textsc{aeasgd}. This is due to an equilibrium condition which is present in \textsc{easgd}~\cite{Hermans:2276711}. Since we increase the amount of asynchrony in the optimization procedure, workers will reach the equilibrium condition faster because the elastic difference is computed based on the most recent parameterization of the central variable. Meaning, as soon as \textsc{aeasgd} is done computing $\lambda$ iterations, the central variable is pulled to the worker where the elastic difference is computed based on the recently pulled central variable, which is very stale due to the low communication frequency and high number of asynchronous workers. As a result, \textsc{aeasgd} is rather slow reaching a better training accuracy in the presence of small gradients compared to other optimization algorithms.

\begin{figure}
  \centering
  \subfigure[$\mathcal{E} = 0.964$]{\includegraphics[width=.31\linewidth]{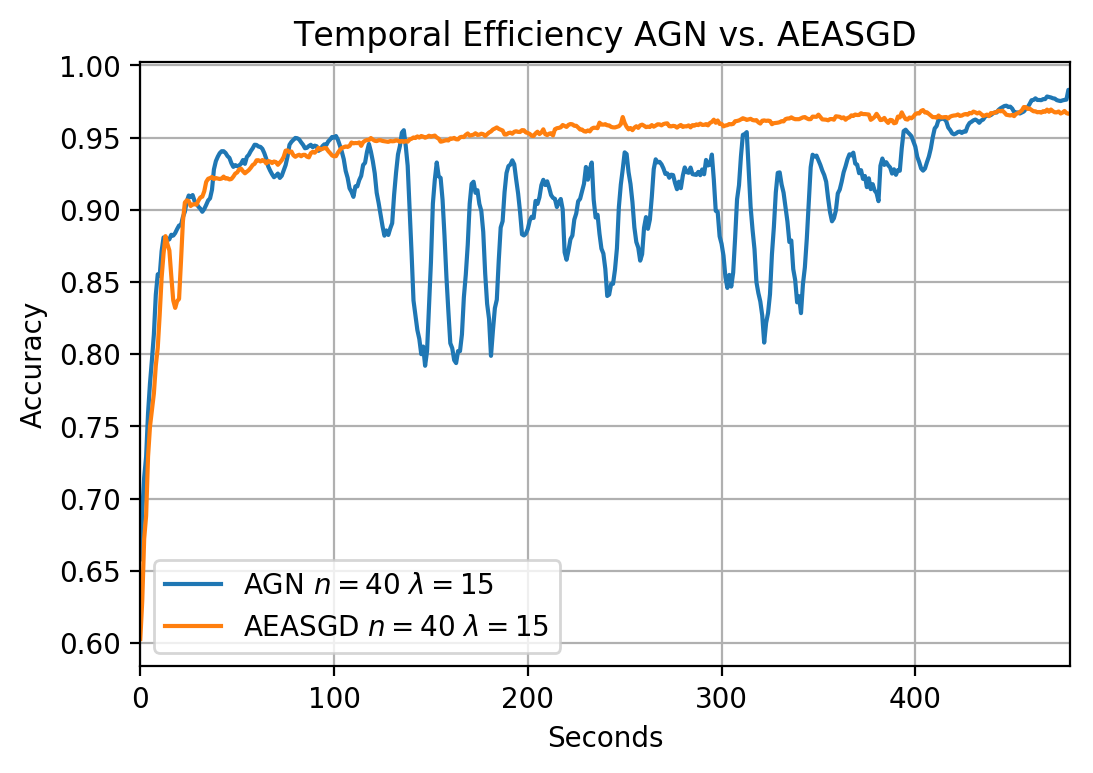}}
  \subfigure[$\mathcal{E} = 0.999$]{\includegraphics[width=.31\linewidth]{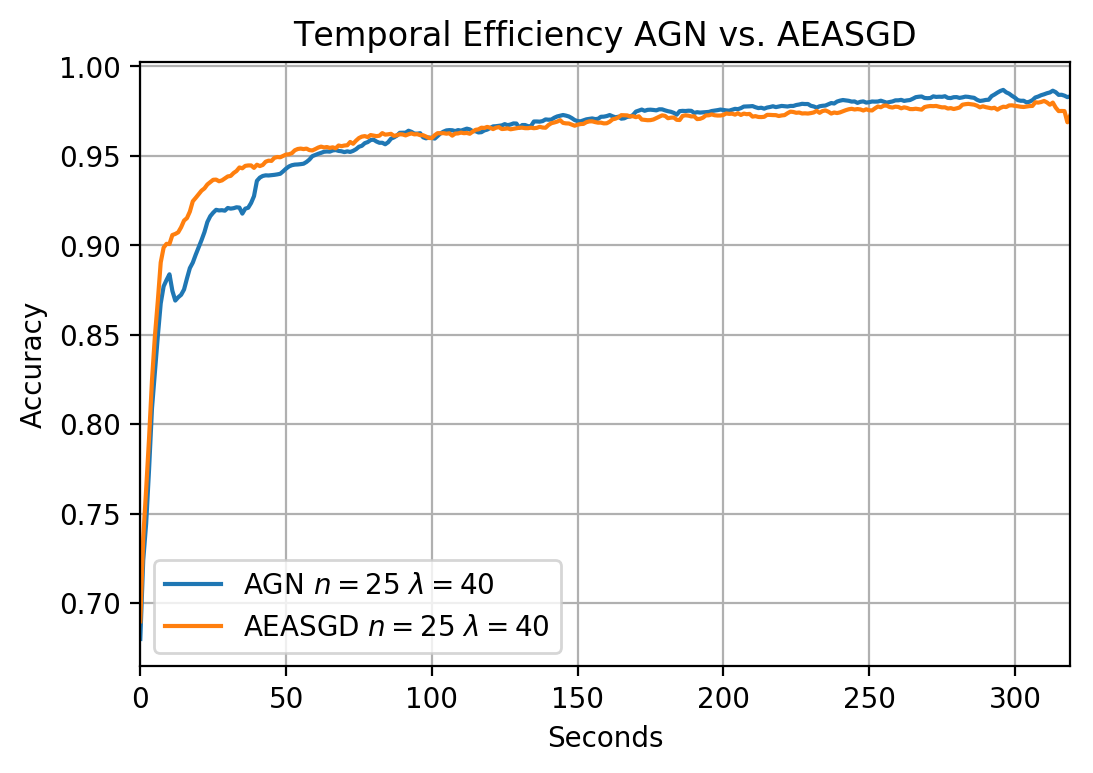}}
  \subfigure[$\mathcal{E} = 1.044$]{\includegraphics[width=.31\linewidth]{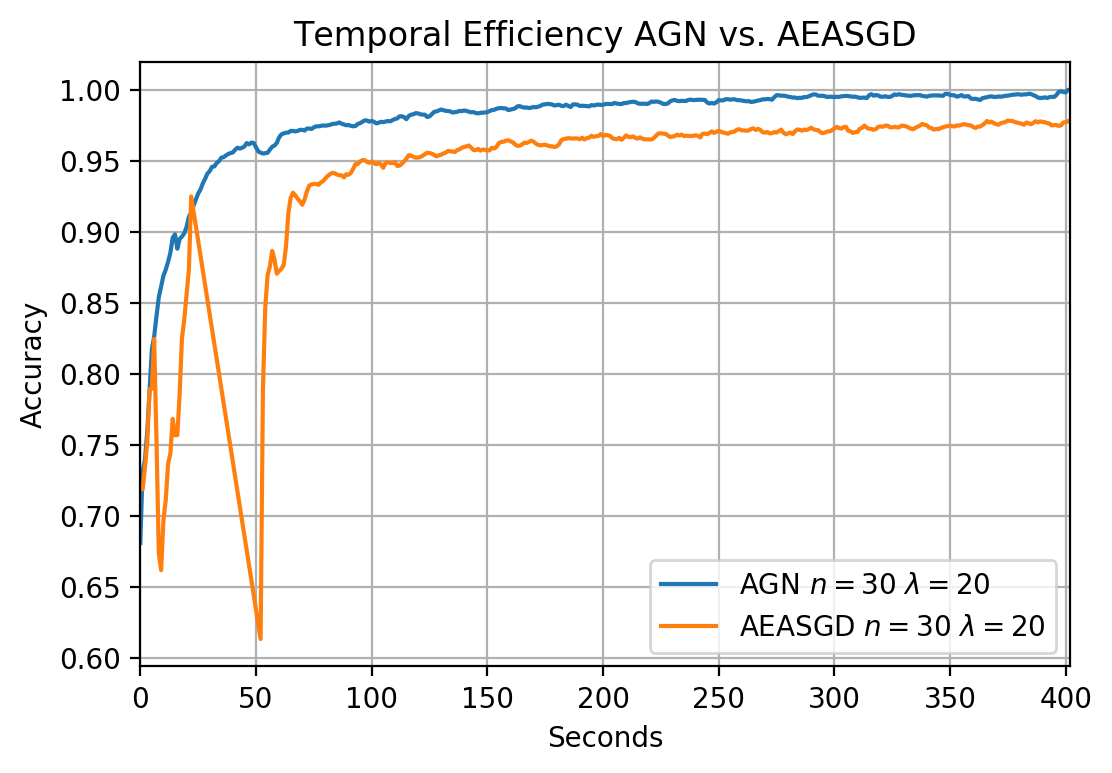}}
  \caption{Several accuracy plots of \textsc{agn} and \textsc{aeasgd}. All subfigures show the computed \emph{temporal efficiency} of \textsc{agn}, which were obtained by applying Equation~\ref{eq:temporal_efficiency}.}
\end{figure}

\begin{table}
  \centering
  \begin{tabular}{|c|c|c|c|c|c|c|c|}
    \hline
    $n$ & $\lambda$ & \textsc{AGN} $t$ & \textsc{AGN} Acc. & \textsc{AEASGD} $t$ & \textsc{AEASGD} Acc. & $\mathcal{E}(\textsc{agn},\textsc{aeasgd})$ \\
    \hline
    \hline
10 & 10 & 1066.08s & \textbf{100.00\%} & \textbf{953.61s} & 99.22\% & \textbf{1.009} \\
\hline
10 & 15 & 864.01s & \textbf{99.53\%} & \textbf{846.86s} & 99.38\% & \textbf{1.012} \\
\hline
10 & 20 & 886.34s & \textbf{99.38\%} & \textbf{804.07s} & 98.91\% & \textbf{1.003} \\
\hline
10 & 25 & 855.51s & 98.75\% & \textbf{784.46s} & \textbf{98.91\%} & 0.999 \\
\hline
10 & 30 & \textbf{886.01s} & \textbf{99.22\%} & 930.73s & 98.91\% & 0.988 \\
\hline
10 & 35 & 850.87s & 98.44\% & \textbf{798.74s} & \textbf{99.22\%} & 0.990 \\
\hline
10 & 40 & 845.21s & \textbf{98.75\%} & \textbf{791.04s} & 97.66\% & 0.990 \\
\hline
20 & 10 & 839.94s & \textbf{99.26\%} & \textbf{821.12s} & 97.90\% &  0.983 \\
\hline
20 & 15 & \textbf{571.17s} & \textbf{100.00\%} & 610.88s & 98.52\% & \textbf{1.018} \\
\hline
20 & 20 & \textbf{432.93s} & \textbf{99.38\%} & 510.72s & 97.78\% & \textbf{1.022} \\
\hline
20 & 25 & 479.72s & \textbf{99.63\%} & \textbf{421.50s} & 97.86\%  & \textbf{1.009} \\
\hline
20 & 30 & 433.36s & \textbf{99.42\%} & \textbf{429.16s} & 98.36\% & \textbf{1.007} \\
\hline
20 & 35 & 418.83s & \textbf{98.52\%} & \textbf{409.86s} & 98.19\% & \textbf{1.002} \\
\hline
20 & 40 & 434.86s & \textbf{98.44\%} & \textbf{420.46s} & 97.66\% & 0.997 \\
\hline
40 & 10 & \textbf{748.94s} & 94.17\% & 1256.09s & \textbf{96.57\%} & \textbf{1.044}  \\
\hline
40 & 15 & \textbf{506.25s} & 95.99\% & 534.42s & \textbf{96.88\%} & 0.964 \\
\hline
40 & 20 & \textbf{383.51s} & \textbf{99.24\%} & 412.37s & 96.65\% & \textbf{1.027} \\
\hline
40 & 25 & \textbf{308.15s} & \textbf{98.86\%} & 347.50s & 96.65\% & \textbf{1.025}  \\
\hline
40 & 30 & 351.54s & \textbf{98.66\%} & \textbf{305.50s} & 96.47\% & 0.997 \\
\hline
40 & 35 & 279.30s & \textbf{98.73\%} & \textbf{252.70s} & 96.32\% & \textbf{1.009} \\
\hline
40 & 40 & 257.62s & \textbf{97.88\%} & \textbf{250.74s} & 96.65\% & \textbf{1.009} \\
\hline
  \end{tabular}
  \caption{Summary of \textsc{agn} and \textsc{aeasgd} experiments using different distributed hyperparameters ($n$ and $\lambda$). From these experiments we find that \textsc{agn} performs better in terms of training and validation accuracy in the presence of a higher number of asynchronous workers, and a reduced communication frequency. We also include the temporal efficiency of \textsc{agn} and \textsc{aeasgd} compared to different distributed hyperparameters. Using this information, we can deduce that \textsc{agn} is outperforming \textsc{aeasgd} in 69.73\% of the cases, which is significantly better. Furthermore, this statistic includes cases which are known where \textsc{agn} is performing badly, i.e., small amount of asynchrony, low communication frequency, and high amount of asynchrony, and high communication frequency.}
  \label{table:agn_experiments_summary}
\end{table}

\begin{figure}
  \centering
  \subfigure[$\lambda = 10$]{\includegraphics[width=.3\linewidth]{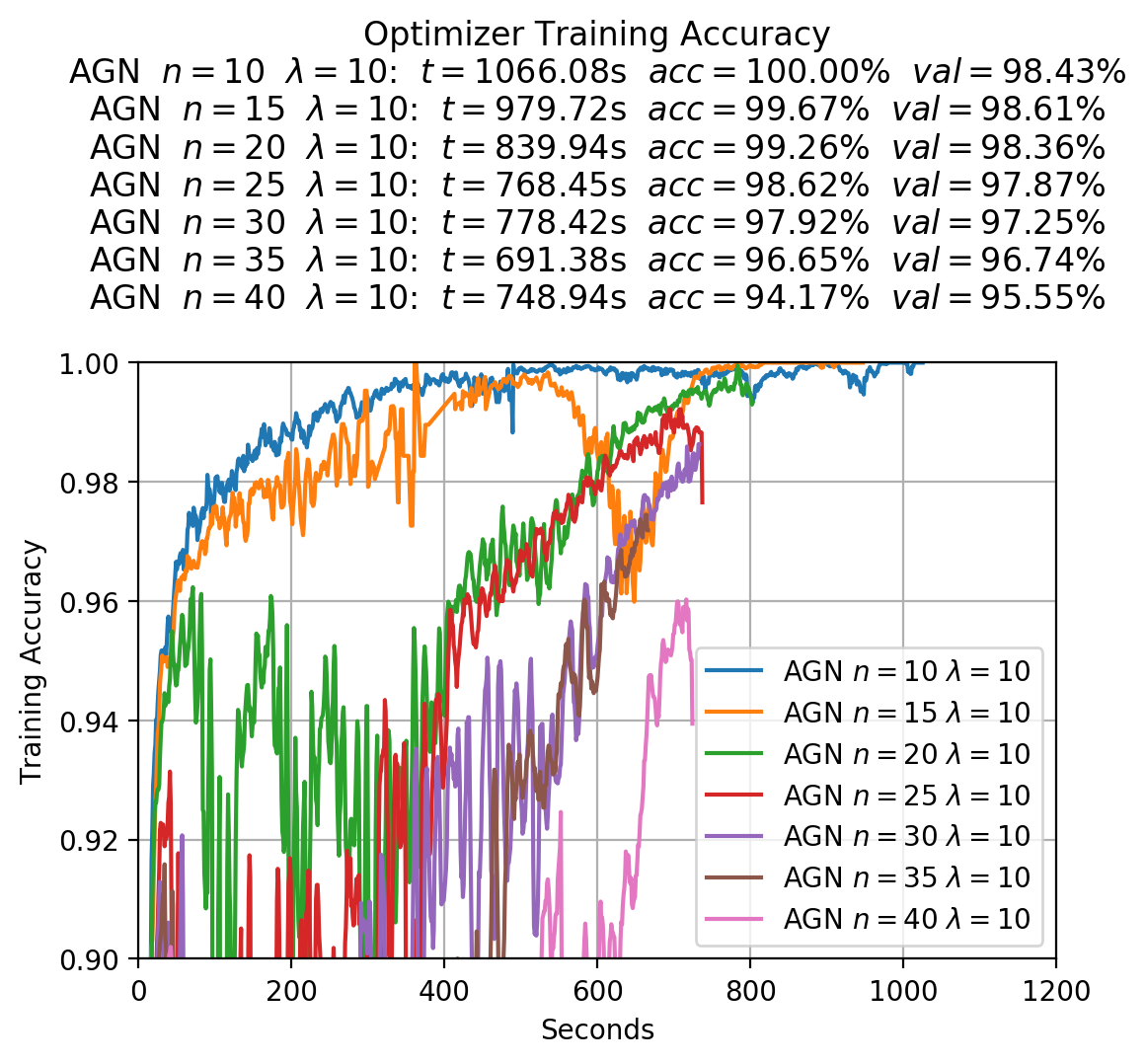}}
  \subfigure[$\lambda = 15$]{\includegraphics[width=.3\linewidth]{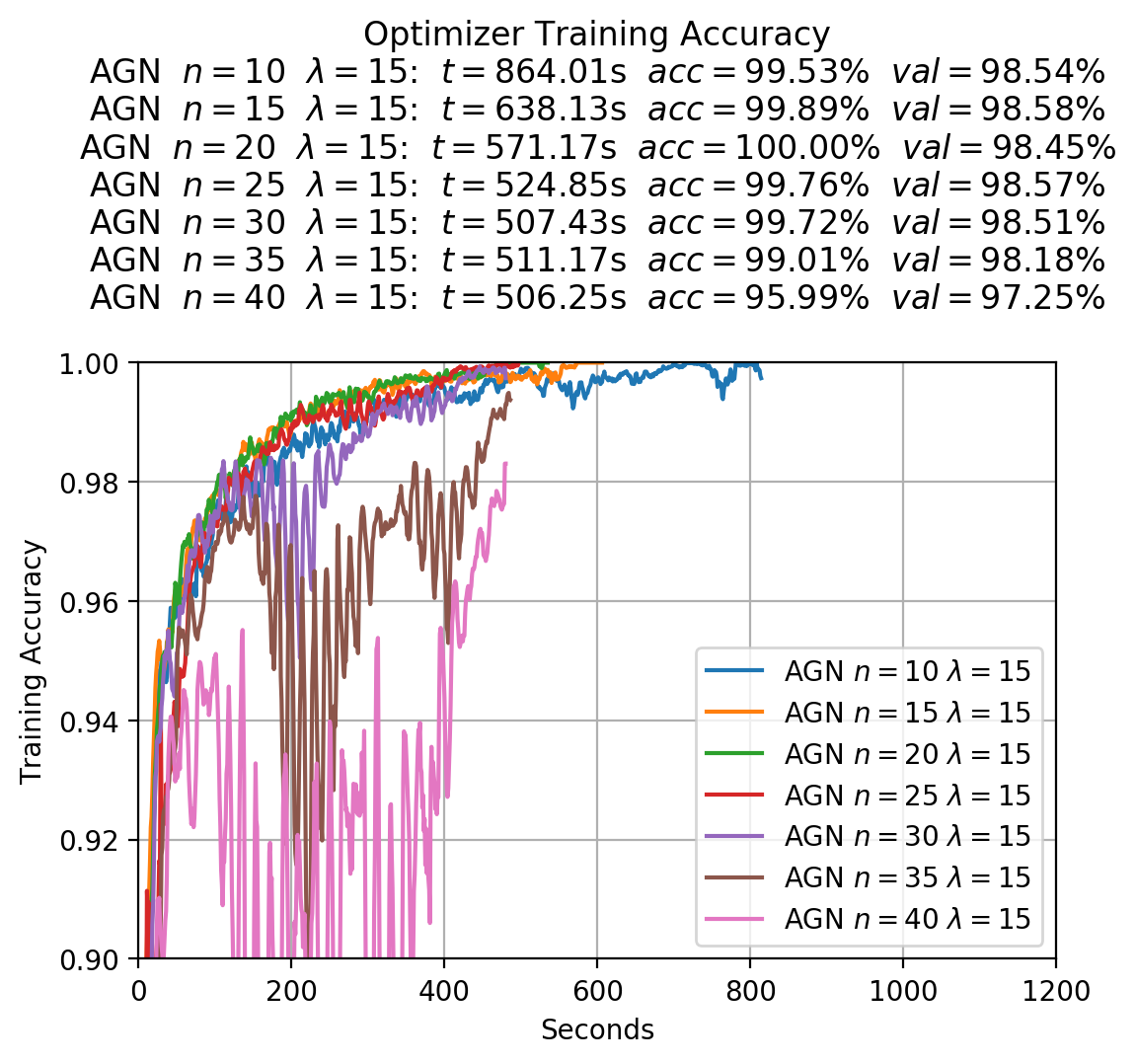}}
  \subfigure[$\lambda = 20$]{\includegraphics[width=.3\linewidth]{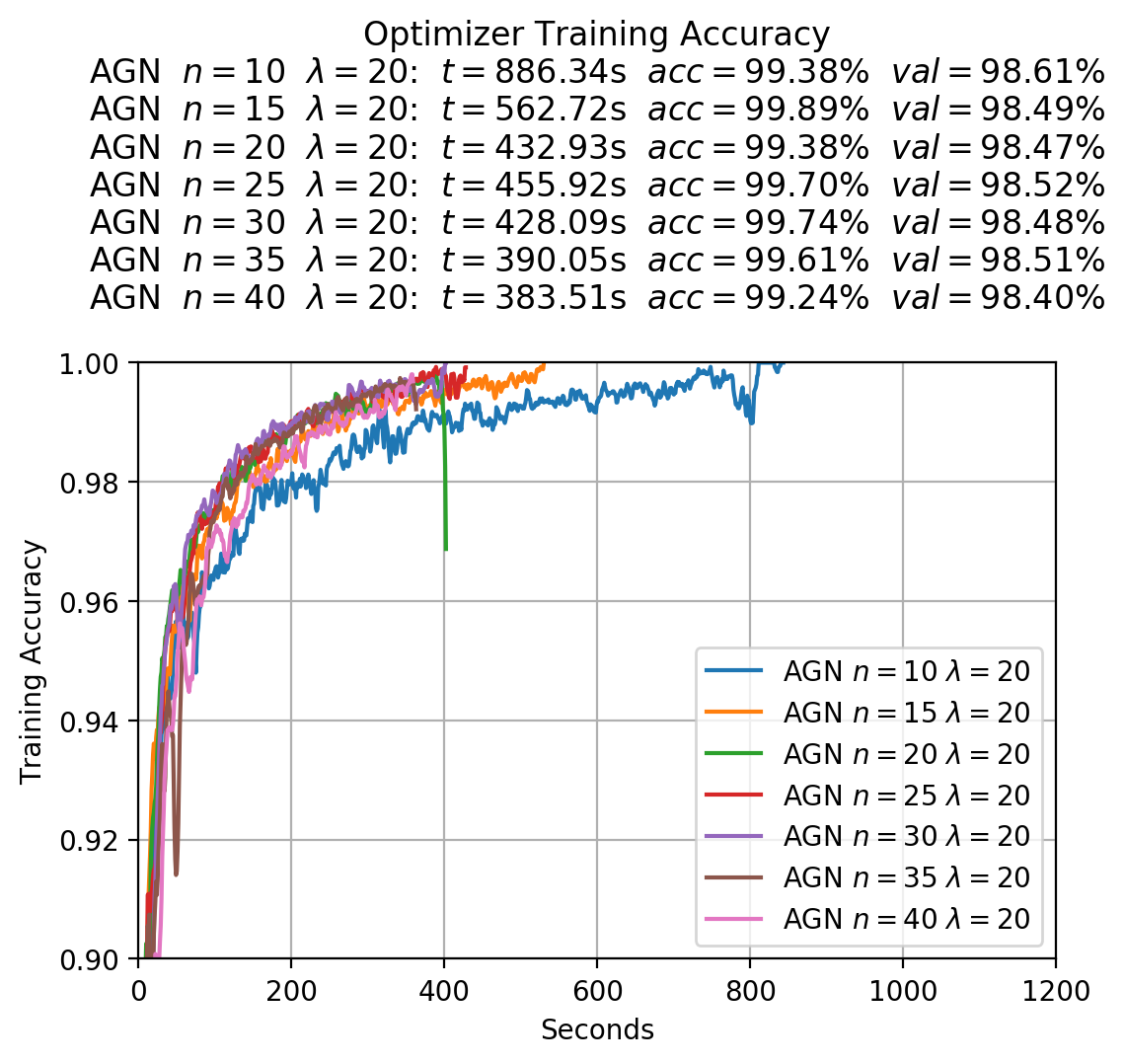}}
  \subfigure[$\lambda = 25$]{\includegraphics[width=.3\linewidth]{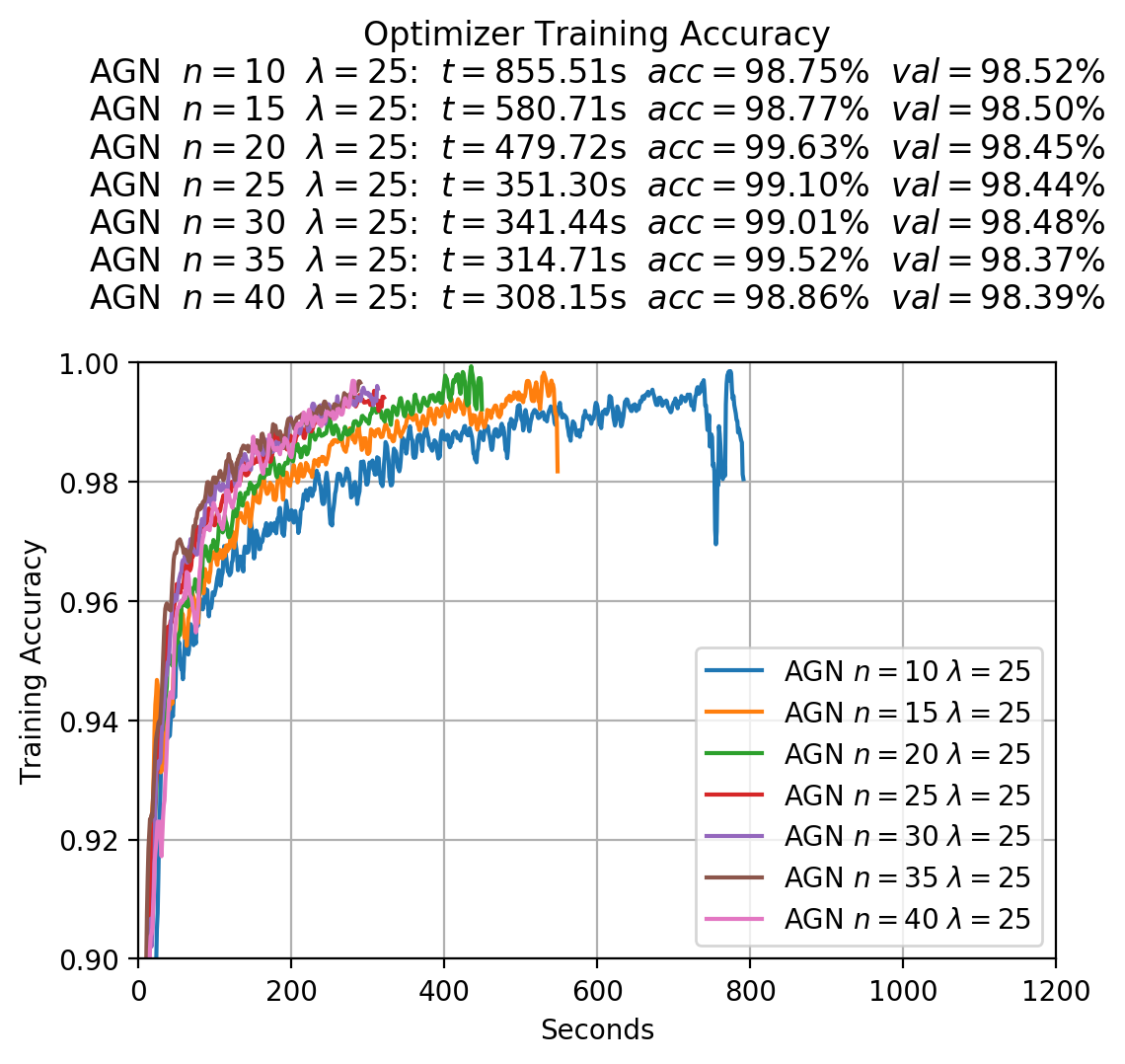}}
  \subfigure[$\lambda = 30$]{\includegraphics[width=.3\linewidth]{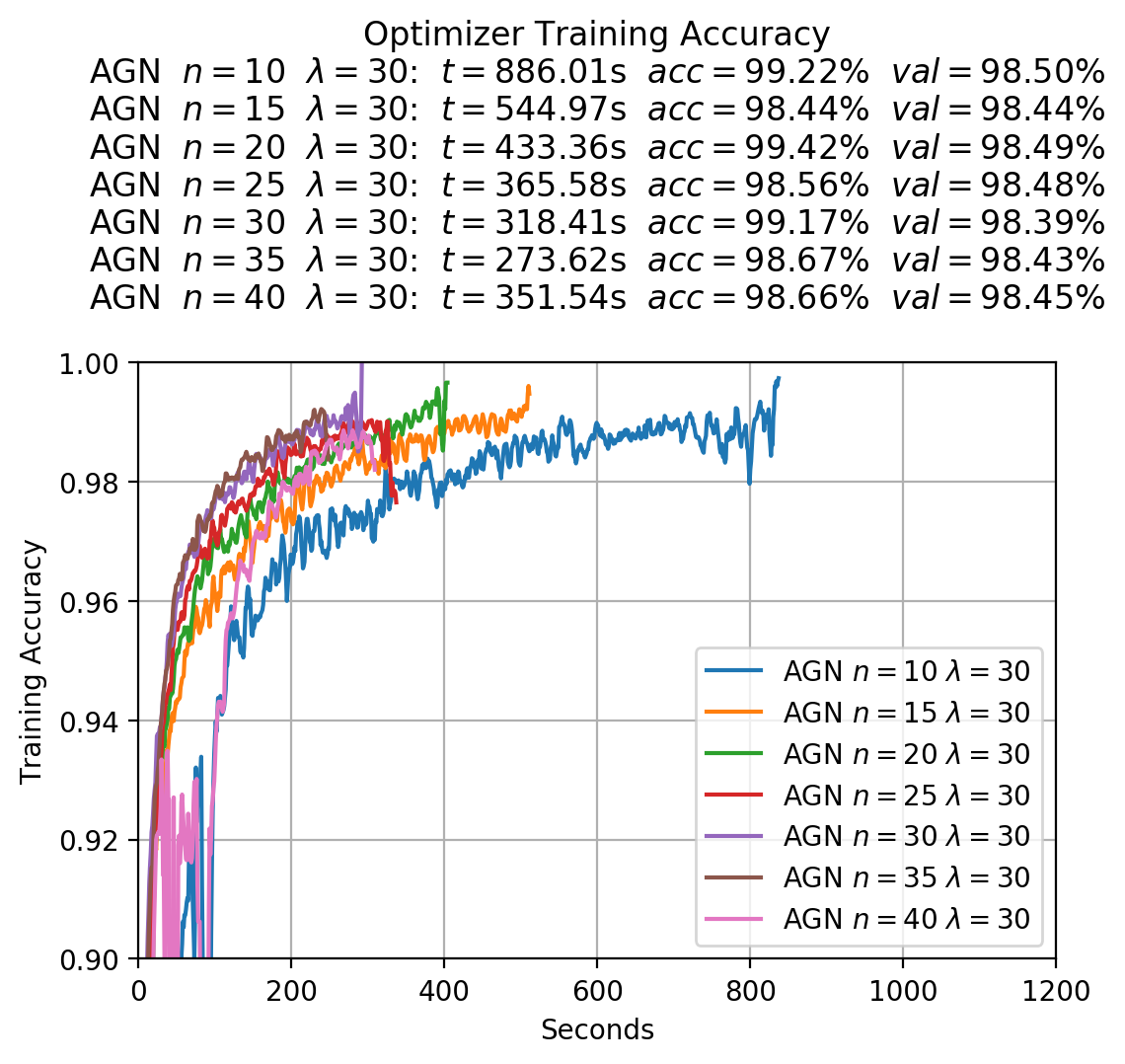}}
  \subfigure[$\lambda = 40$]{\includegraphics[width=.3\linewidth]{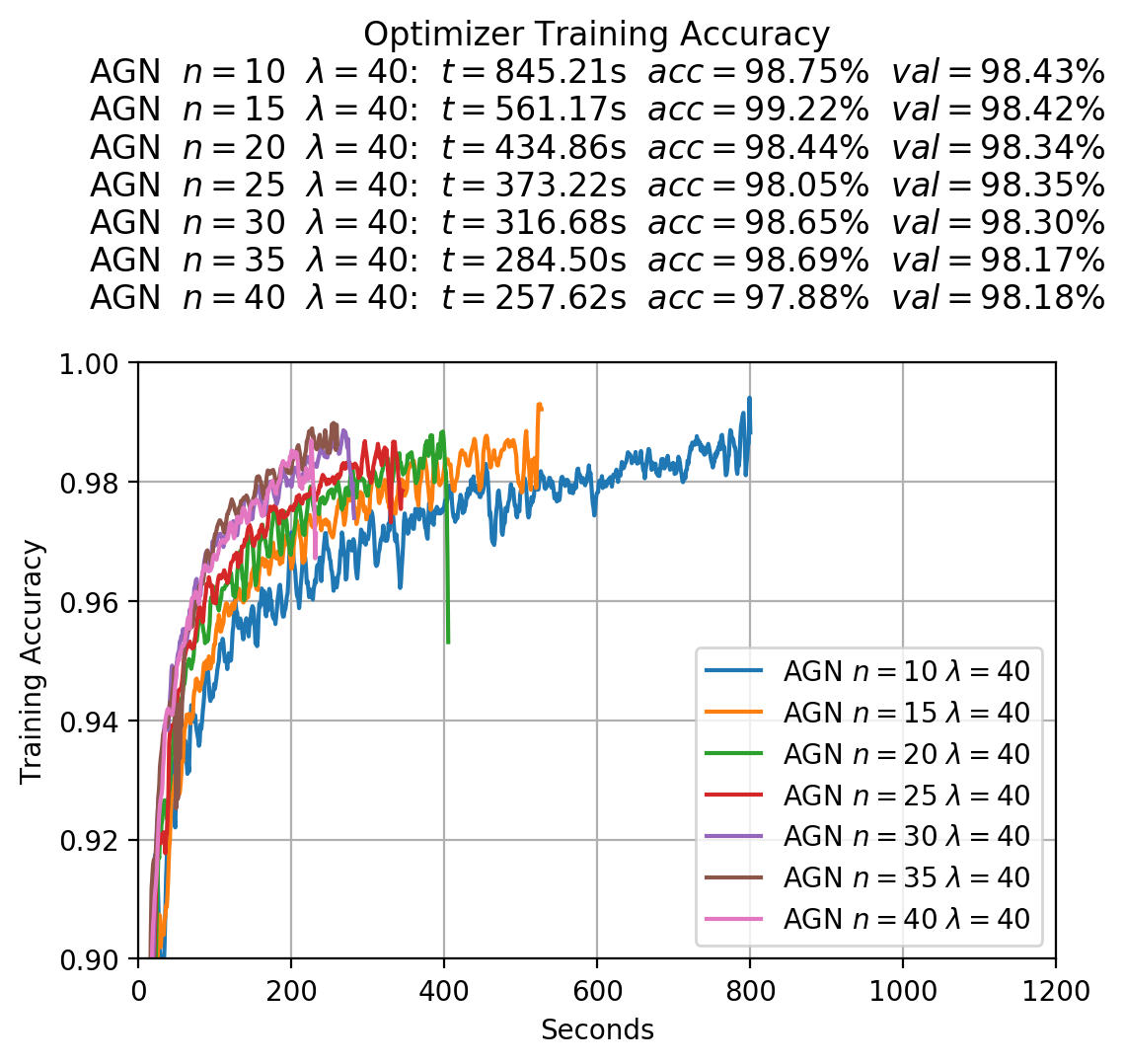}}
  \caption{In this experiment we clamp the communication frequency, but vary the number of asynchronous workers. Due to the equal communication frequency, we can observe good scaling properties of \textsc{agn}. In most cases doubling the number of workers, reducing the training time by half and is more temporally efficient. However, for larger number of workers $n > 30$ we do not observe a reduction of training time. This is due to the implementation of the parameter server used in our experiments, which is based on Python threads instead of Python processes. Furthermore, note that reducing the amount of computational resources might actually benefit the training accuracy of the central variable, as a smaller number of asynchronous workers reduces the amount of staleness that can be Incorporated in the central variable.}
  \label{fig:agn_experiments_lambdas}
\end{figure}

\section{Conclusion}
\label{sec:conclusion}

This work introduces a novel distributed optimization procedure. We make identical \emph{practical} assumptions with respect to constraints as \textsc{easgd}, i.e., high communication costs. However, \textsc{agn} is not effected by equilibrium conditions which incapacitate the converge rate of the optimization process, which are present in \textsc{easgd}~\cite{Hermans:2276711}. As a result, we turned to \textsc{downpour}, and allowed for more local exploration by sending \emph{accumulated gradients} to the parameter server. However, this approach diverged even faster than regular \textsc{downpour}, with the difference that data was processed significantly faster due to the reduced waits.\\

Therefore, \textsc{downpour} was adapted to use the time between parameter updates more efficiently by computing a \emph{better} gradient based on a normalized sequence of first-order gradients, thus obtaining Accumulated Gradient Normalization. Furthermore, we show that \textsc{agn} outperforms existing distributed optimizers in terms of convergence rate in the presence of a large amount of concurrency and communication constraints. Since \emph{stability} is also important in distributed optimization, we introduce a new metric called \emph{temporal efficiency} which is defined as the ratio between the integrated area of training metrics of two different optimizers. As a result, not only the final training accuracy is considered, but also the stability, and time required to reach an accuracy level.\\

To conclude, \textsc{agn} achieves this result by computing a better \emph{direction} of the gradient based on a sequence of first order gradients which can be computed locally without any communication with the centralized model. This direction also reduces the negative effects of implicit momentum in highly concurrent environment as most workers will point in the general direction of a minimum, thereby making \textsc{agn} more robust to distributed hyperparameterization such as the number of workers and communication frequency. Although, as the number of workers increases, a decline in training and validation accuracy is still observed due to the amount of implicit momentum. A natural response would be to lower the communication frequency (increase $\lambda$). However, this would increase the amount of local exploration with the possibility that a subset of workers might end up in different minima.\\

For possible future work, it would be of interest to explore if adaptive communication frequencies might benefit the optimization process. In fact, in small gradient environments (e.g., close to a minima) it might be beneficial to \emph{not} normalize the accumulated gradients since the first-order gradient updates are relatively small anyway.

\section{Acknowledgements}
\label{sec:acknowledgements}

We would like the team members of CERN IT-DB for supporting this work, especially Zbigniew Baranowski and Luca Canali. We also thank Vlodimir Begy for the insightful discussions. Finally, we thank the Dutch National Science Foundation NWO for financial support through the project SWARMPORT (NWO project number 439.16.108).

\newpage
\bibliography{acml17}


\end{document}